\documentclass[letterpaper,10 pt,conference]{ieeeconf}
\IEEEoverridecommandlockouts
\usepackage{graphics} 
\usepackage{epsfig}   
\usepackage{mathptmx} 
\usepackage{amsmath}  
\usepackage{amssymb}  

\usepackage{algorithm}
\usepackage{algpseudocode}
\usepackage{bm}
\usepackage{enumerate}
\usepackage{graphicx}
\usepackage{subfig}

\graphicspath{{images/}}

\makeatletter
\renewcommand{\fnum@figure}{Fig. \thefigure}
\makeatother

\newcommand{\RR}{\mathbb{R}}

\newcommand{\etal}{\emph{et al.}~}

\title{\LARGE \bf Intuitive Robot Integration via Virtual Reality Workspaces}

\author{Minh Q.~Tram, Joseph M.~Cloud, and William J.~Beksi
\thanks{The authors are with the Department of Computer Science and
        Engineering, The University of Texas at Arlington, Arlington, TX, USA.
        Emails:
        minh.tram@mavs.uta.edu,
        joe.cloud@mavs.uta.edu,
        william.beksi@uta.edu.}}

\begin{document}

\maketitle
\pagestyle{plain}

%
%

\begin{abstract}
As robots become increasingly prominent in diverse industrial settings, the
desire for an accessible and reliable system has correspondingly increased. Yet,
the task of meaningfully assessing the feasibility of introducing a new robotic
component, or adding more robots into an existing infrastructure, remains a
challenge. This is due to both the logistics of acquiring a robot and the need
for expert knowledge in setting it up. In this paper, we address these concerns
by developing a purely virtual simulation of a robotic system. Our proposed
framework enables natural human-robot interaction through a visually immersive
representation of the workspace. The main advantages of our approach are the
following: (i) independence from a physical system, (ii) flexibility in defining
the workspace and robotic tasks, and (iii) an intuitive interaction between the
operator and the simulated environment. Not only does our system provide an
enhanced understanding of 3D space to the operator, but it also encourages a
hands-on way to perform robot programming. We evaluate the effectiveness of our
method in applying novel automation assignments by training a robot in virtual
reality and then executing the task on a real robot.
\end{abstract}

\begin{keywords}
Virtual Reality and Interfaces; Human-Centered Automation; Human-Robot
Collaboration
\end{keywords}

\section{Introduction}
\label{sec:introduction}
Robots have become the cornerstone of many industrial operations due to their
efficiency, productivity, and reliability. The dependency on robots is
ultimately rooted in the need for an autonomous workforce that can achieve high
throughput with low downtime in order to meet increasing production demands.
With applications spread across large industries (e.g., agricultural operations,
automotive manufacturing, pharmaceutical packaging, etc.) robots have become
irreplaceable and serve a critical role in the supply chain workflow. This
aspect is further emphasized during times of national emergency such as the
recent COVID-19 pandemic \cite{shen2021covid}. Thus, the demand for more robust
and highly-integrated robotic systems increases as the industrial sector
intensifies its growth in automation.

\begin{figure}[ht]
\centering
\includegraphics[width=\linewidth, height=4.7in]{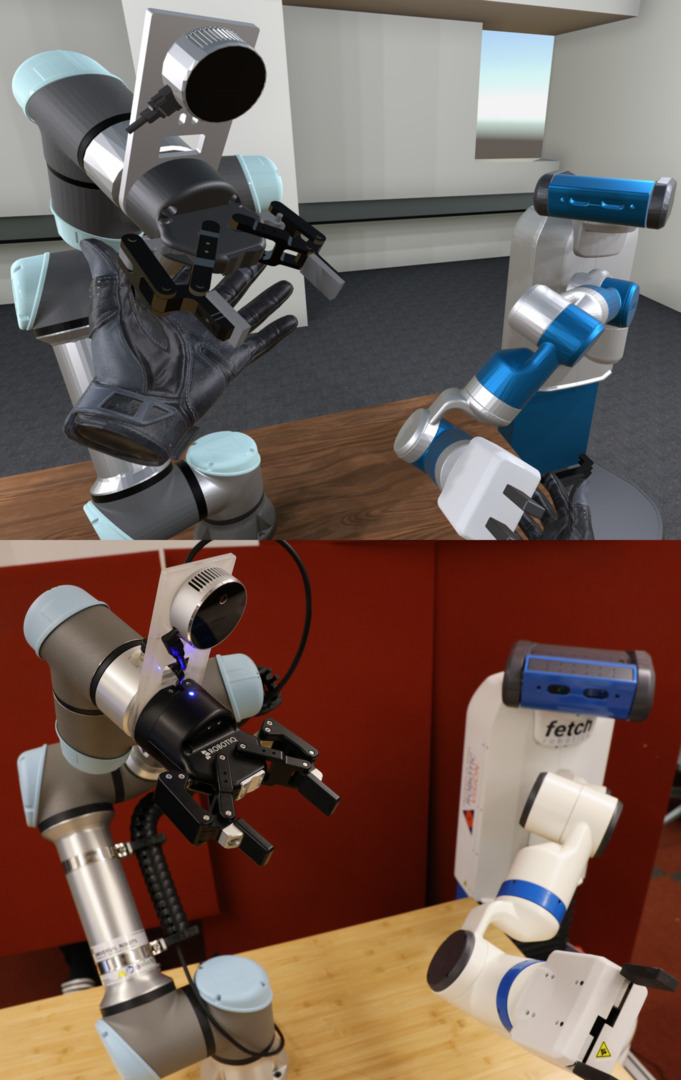}
\caption{A sample scene captured in our VR workspace from the point of view of
the operator. The operator can directly interact with individual robot joints
simply by grabbing or pushing them as they normally would in a real
environment.}
\label{fig:concept_overview}
\end{figure}

Although the deployment of robots continues to rise, the task of determining the
feasibility of using a new robotic component, either in a novel work environment
or adding more robots to collaborate within an existing infrastructure, remains
extremely difficult \cite{sanneman2021state}. Acquiring and implementing capable
industrial robots is often undesirable for small and medium-sized enterprises.
This is due to high initial investment costs and a nontrivial integration
process that requires expert knowledge with no guarantee of profitable returns.
Meanwhile, the expanding virtual, augmented, and mixed reality (VAMR) industry
has yielded incredibly useful tools and techniques to enhance the robotic
development process.

VAMR provides an \textit{intuitive understanding} of 3D space while
simultaneously providing \textit{additional information} to the operator. This
leads to an overall \textit{enhanced} operator experience and
\textit{encourages} a hands-on approach to robot programming
\cite{arevalo2020cues}. Nonetheless, VAMR-based human-robot interaction (HRI)
research is generally focused on teleoperation for task-specific applications
\cite{gharaybeh2019teleop, wang2021weld}, human-in-the-loop digital twins
\cite{tsokalo2019twins, puljiz2019hand}, or machine learning with densely
overlayed information interfaces \cite{zhang2018learning,liu2018learning,
dyrstad2018teaching}. However, these frameworks have the following
\textit{disadvantages}: (i) an assumption of a pre-installed, functioning
robotic system, (ii) a lack of support for high-dexterity interaction, and (iii)
a focus on single-robot use cases.

In this work, we propose a simulated robotic environment capable of representing
and integrating operator-configurable workspaces for \textit{multiple}
collaborative robots, Fig.~\ref{fig:concept_overview}. The primary objective of
this research is to create a virtual reality (VR) workspace that allows an
operator to \textit{immersively explore} and \textit{efficiently assess} the
integration of a desired robotic system into a physical environment. We also aim
to make this work a \textit{foundation} for future robot programming and
operator training. In summary, our contributions are the following.
\begin{itemize}
  \item We eliminate the dependency on a physical robotic system by moving
  visualization and interaction into a purely virtual environment.
  \item We simulate multiple collaborative robots in realistic setup and usage
  scenarios (e.g., direct imitation learning, learning from demonstration).
  \item We provide operators with an intuitive and informative representation
  and control of virtual robots.
\end{itemize}
The source code and Docker image associated with this project are publicly
available at \cite{vamr2023}.

The remainder of the paper is organized as follows. We provide an overview of
related research in Section~\ref{sec:related_work}. The details of our system
are presented in Section~\ref{sec:virtual_reality_robotic_workspace}. Our
evaluation use cases and results are discussed in Section~\ref{sec:evaluation}.
In Section~\ref{sec:conclusion_and_future_work}, we conclude and provide
directions for future work.

\section{Related Work}
\label{sec:related_work}
The practice of leveraging VAMR to enhance the experience of robot programming,
and provide interaction and information through either see-through displays or
head-mounted displays (HMD), is very popular. However, the use of augmented or
mixed reality far exceeds that of VR. Surveys have shown that for effective HRI,
it is crucial to provide sufficient information not only to the robot, but also
to the operator \cite{mukherjee2022survey}. To this extent, the following common
setup is favorable for many researchers: (i) an overlaying interface, provided
by VAMR, communicating with real robot hardware through a Robot Operating System
(ROS) \cite{quigley2009ros} backend; (ii) motion planning delegated to either
their own implementation of forward kinematics (FK) and inverse kinematics (IK),
or MoveIt \cite{coleman2014reducing}, a well-known robot motion planning
framework.

Visualizing a planned robot action, before it executes, plays an important role
in a collaborative workspace in terms of human safety and correct execution.
Quintero \etal \cite{quintero2018programming} highlighted the advantages of
having additional information and visualization for robot motion planning, while
remaining relatively free from the use of any handheld devices, by facilitated
on-board hand tracking of their HMD using the Microsoft HoloLens. They also
explored the use of multiple instruction input sources, such as speech and
gesture, as an alternative means of robot control. Perez \etal
\cite{perez2019virtual} demonstrated a different, more VR-oriented approach,
where a warehouse was projected onto the operator's point of view by the use of
3D scanners and scene mapping through post-processing via third-party software.
Control of their robot was rudimentary in that it indirectly communicated to the
manufacturer's interface through a VR headset rather than directly manipulating
the scene.

In general, related work on this topic has one or more of the following
objectives: (i) programming or operating a robot through VAMR
\cite{lambrecht2012spatial,quintero2018programming,kot2018hololens,gadre2019mr,
ostanin2020mr}, or (ii) leveraging VAMR as a foundation for other purposes such
as imitation learning \cite{zhang2018learning, dyrstad2018teaching}. There has
also been significant effort made towards the use of VAMR to analyze complex
behaviors of multiple collaborative robots (e.g., if they can accomplish a
common goal with human partners rather than working around them). These works
often target mobile rather than stationary robots \cite{ghiringhelli2014multi,
chen2019mobile}, and they typically use augmented or mixed reality. In contrast,
our work is focused solely on utilizing VR as a simulation, training, and
integration analysis framework without the need for physical robots.
Nevertheless, our system is still capable of transferring execution over to a
real platform. Moreover, we prioritize direct hands-to-object interactions
rather than alternative means of control. This allows us to emulate ultra
realistic interplay between the operator and the robots.

\section{Virtual Reality Robotic Workspace}
\label{sec:virtual_reality_robotic_workspace}
In this section, we present our virtual reality robotic workspace (VRRW)
architecture and how each component facilitates the proposed framework.

\subsection{System Overview and Design Choices}
\label{subsec:overview}

\begin{figure}
\includegraphics[width=\linewidth]{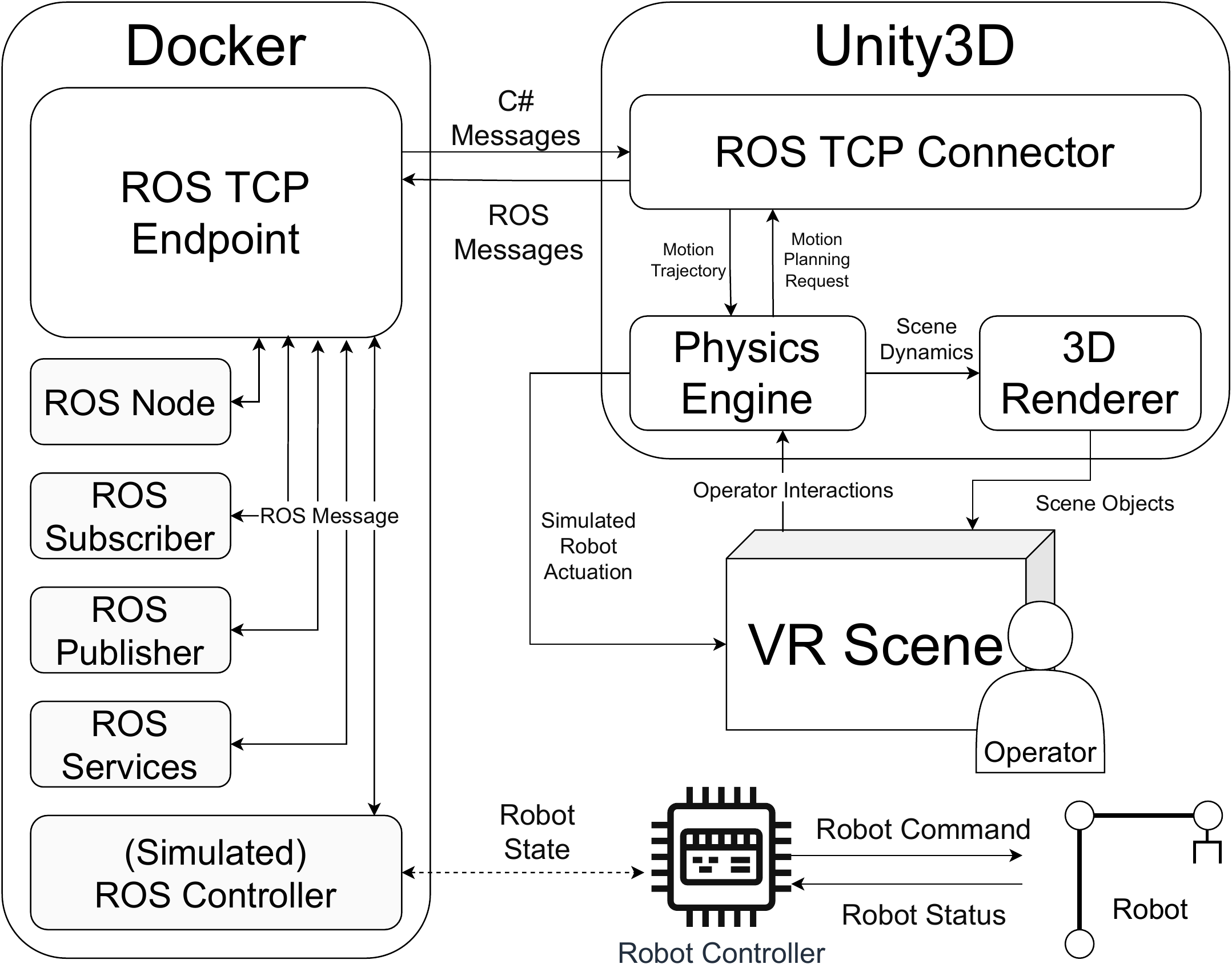}
\caption{An overview of the system components including the underlying
communication and rendering pipelines.}
\label{fig:system_components_overview}
\end{figure}

Our VRRW is comprised of the following three major components: (i) VR HMD, (ii)
VR rendering software, and (iii) robotic simulation and planning backend. This
section describes each component and the functionality it provides for our
proposed approach. Fig.~\ref{fig:system_components_overview} details the
communication pipeline across the hardware and software components. 

For the VR headset, we developed our system using the Steam Valve Index
\cite{valveindex}. The choice of headset is dependent on the display resolution
and refresh rates. This is a vital point of consideration since we aim to reduce
the effects of initial real-to-sim and sim-to-real transitions such as motion
sickness and disorientation. A high-resolution display ($1440 \times 1600$
pixels per eye), high refresh rate (up to 144 Hz), low pixel illumination
periods (0.330 ms to 0.530 ms), and accurate live tracking of the headset
position all contribute to reducing these VR-related issues.

We utilized the Unity \cite{unity} game engine as our environment rendering
software. In recent years, due to an increased interest in VAMR for robotics,
Unity development has branched off into providing a software suite capable of
rendering and representing robots and robotic environments. This software suite,
commonly known as the Unity Robotics
Hub\footnote{https://github.com/Unity-Technologies/Unity-Robotics-Hub}, not only
provides proper two-way communication between Unity and ROS, but it also
contributes to how robots are represented within Unity itself. 

For robot simulation and motion planning, we made use of ROS 1 Noetic running on
Ubuntu 20.04 from a Docker image. We loosely use the term \textit{simulation}
since we do not run a full-scale Gazebo \cite{koenig2004design} simulation where
the simulated robot is practically indistinguishable from a real setup, nor are
we communicating with the official simulation backend. Instead, ROS serves the
following three purposes: (i) communication, (ii) joint-state correspondence,
and (iii) motion planning through MoveIt.

\subsection{Unity 3D Game Engine}
\label{subsec:unity_3d_game_engine}
Unity3D (Unity) was originally a cross-platform game engine targeting developers
rather than VAMR roboticists. As VAMR usage in the gaming industry dramatically
increased, Unity has developed a sophisticated infrastructure to accommodate
high-fidelity games. This includes both visual and interactive support, along
with the addition of more VAMR devices from different manufacturers.
Consequently, VAMR software diverges into other categories beyond just games as
it is now capable enough to render visually appealing and physically accurate
environments. Most prominently, Unity has been used for architecture model
tours, automotive showrooms, cinematic works, and more recently robotic
development.

Unity provides three primary tools for robotic development. The first tool is a
graphic rendering framework capable of supporting VR devices. Not only is Unity
prepackaged with support for in demand, commercially available VR headsets, but
the game engine itself is also very capable of rendering 3D graphics. This
allows us to display the high-fidelity rendering of desired workspaces and
correctly communicate and send frames to the operator's HMD without having the
burden of low-level graphics programming and controller communication.

The second tool is the standardization of ROS communication messages. ROS is
written in C\texttt{++} (or Python) and communicates through TCP via
standardized serializable messages, while Unity is written in C\#. Previously,
this presented difficulties as developers needed to handle network
communication with discrepancies between ROS and Unity messaging standards.
This was done using \textit{ROSBridge} (aka \textit{ROS\#})
\cite{krupke2018multimodal, bambusek2019spatial}. \textit{ROS\#} has been
further developed by Unity into a \textit{ROS TCP Connector} and a \textit{ROS
TCP Endpoint}, and integrated into their software suite as additional packages.

Support for direct importation of the Unified Robot Description Format (URDF) is
the third tool. URDF is a standardized XML format that represents a robot model
along with its articulation. More specifically, Unity will correctly parse and
define the geometry, visual meshes, kinetics, and dynamics attribute of any
robot given its URDF description. Subsequent robot dynamics and kinematics are
handled by PhysX 4.0 \cite{physx}, a full-featured physics engine that Unity
physics is based on.

Internally, each robot joint generated by the URDF description will spawn with
an appropriate articulation type (i.e., fixed, prismatic, revolute, or
spherical) and it will be governed by their respective joint-drive properties.
These properties are defined as
\begin{equation}
  \text{Effect} = \text{Stiffness}\cdot\left(\Delta\text{Position}\right) -
                  \text{Damping}\cdot\left(\Delta\text{Velocity}\right),
\end{equation}
where Effect is the force or torque applied to the joint in any given time
frame, $\Delta\text{Position}$ is the difference between a set joint target and
current joint angle, and $\Delta\text{Velocity}$ is the difference between the
set-joint velocity and current-joint velocity. Thus, if the stiffness is zero,
then the joint becomes a velocity driven joint. Conversely, if damping is zero,
then the joint will only attempt to reach a position driven joint. The forward
and inverse kinematic chain will not be affected since physical interaction and
articulation are handled as a chain defined by the URDF, and Unity respects that
definition.

\subsection{Scene and Robot Description}
\label{subsec:description}

\begin{figure}
\includegraphics[width=\linewidth]{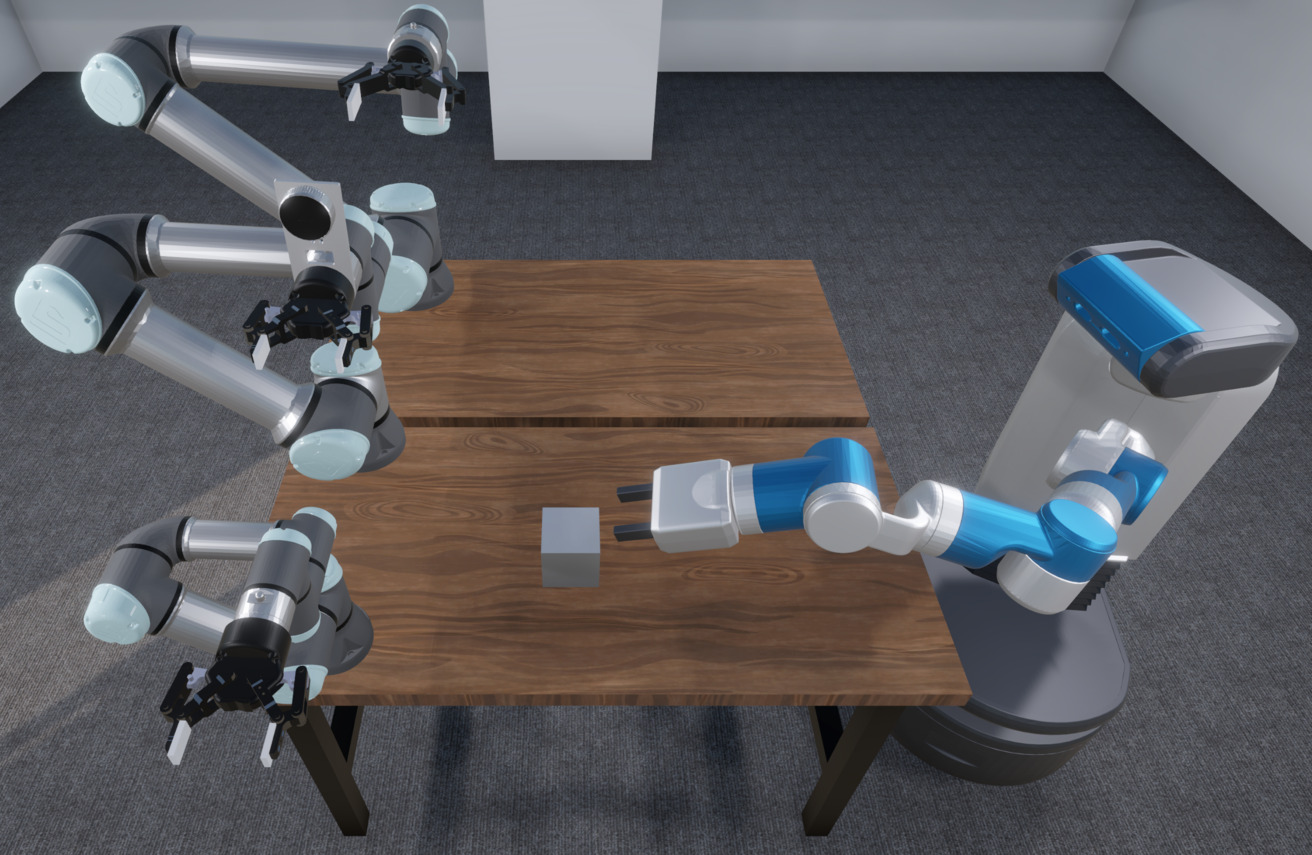}
\caption{A sample virtual workspace with multiple robot variants generated from
URDFs. On the left is a Universal Robots UR3e, UR5e, and UR10e, each equipped
with a different Robotiq 2-Finger (85 mm or 140 mm stroke) end-effector and
sensor. On the right is a Fetch Robotics mobile manipulator. The gray cube on
the table is 10 cm in all dimensions.}
\label{fig:multibot}
\vspace{-1mm}
\end{figure}

While it is desirable to capture as many physical and visual dynamics of a
workspace as possible, such a task is often infeasible due to the infinitely
complicated and chaotic nature of real-world environments. We strive to achieve
a balance between a visually attractive and intuitive workspace, and still retain
a representative physical interaction between the robot and its surrounding
simulated environment. To do this, we present a set of novel solutions for
defining unique workspaces.

Unity, as a 3D game engine, natively supports applicable 3D modeling and
rendering formats. Leveraging this feature, developers can define their
workspace to the level of detail they want with the help of commercial 3D
modeling software. These models can then be manually or programmatically
imported into Unity for visualization. Furthermore, simulated assets can be
redefined and reused on demand to allow flexible customization of an object's
physical and visual characteristics. Alternatively, rather than defining the
entire scene as an exported 3D scene, the developer may also export individual
3D objects (e.g., objects from relevant datasets) and programmatically generate
them on demand into the scene. This is useful for data generation and machine
learning use cases as the base scene can remain static for many iterations,
while the distribution and orientation of the objects can be changed. These
techniques can also be useful for larger scene definitions such as room-scale or
house-scale scenes targeting mobile robots.

Our preferred approach is to define robots in URDF, and then let Unity handle
the 3D generation parsing (Section~\ref{subsec:unity_3d_game_engine}).
Traditionally, to describe a robot in Unity, developers must manually create and
attach all the individual links and joints of the robot while keeping track of
their physical and articulation properties. Hence, two unique, but semantically
identical robot descriptions, are required for Unity to work properly with ROS.
Unity now supports URDF parsing, thus allowing robot descriptions to be directly
imported into the desired simulated scene. Furthermore, ROS supports the use of
XML macros. This allows developers to define their desired robots once and then
subsequently reference that definition in other robot descriptions. As a result,
large and complicated robot descriptions (e.g., numerous robots with multiple
end-effectors) are more achievable as shown in Fig.~\ref{fig:multibot}. 

Lastly, once imported into Unity, a robot model can now be saved directly as an
asset. This further reduces the complexity of introducing additional robots
into a simulated scene. Moreover, it decreases the load time for extensive
robot collaboration projects since Unity does not need to sequentially parse
the URDFs and reconstruct the robot kinematic chain at every launch. Instead, it
just instantiates the complete, pre-generated robot into the scene on the fly.

\subsection{ROS Backend and Motion Planning}
\label{subsec:planning}
\begin{figure}
\centering
\subfloat[Free-drive mode \label{freedrive}]{\includegraphics[width=0.49\linewidth, height=1.6in]{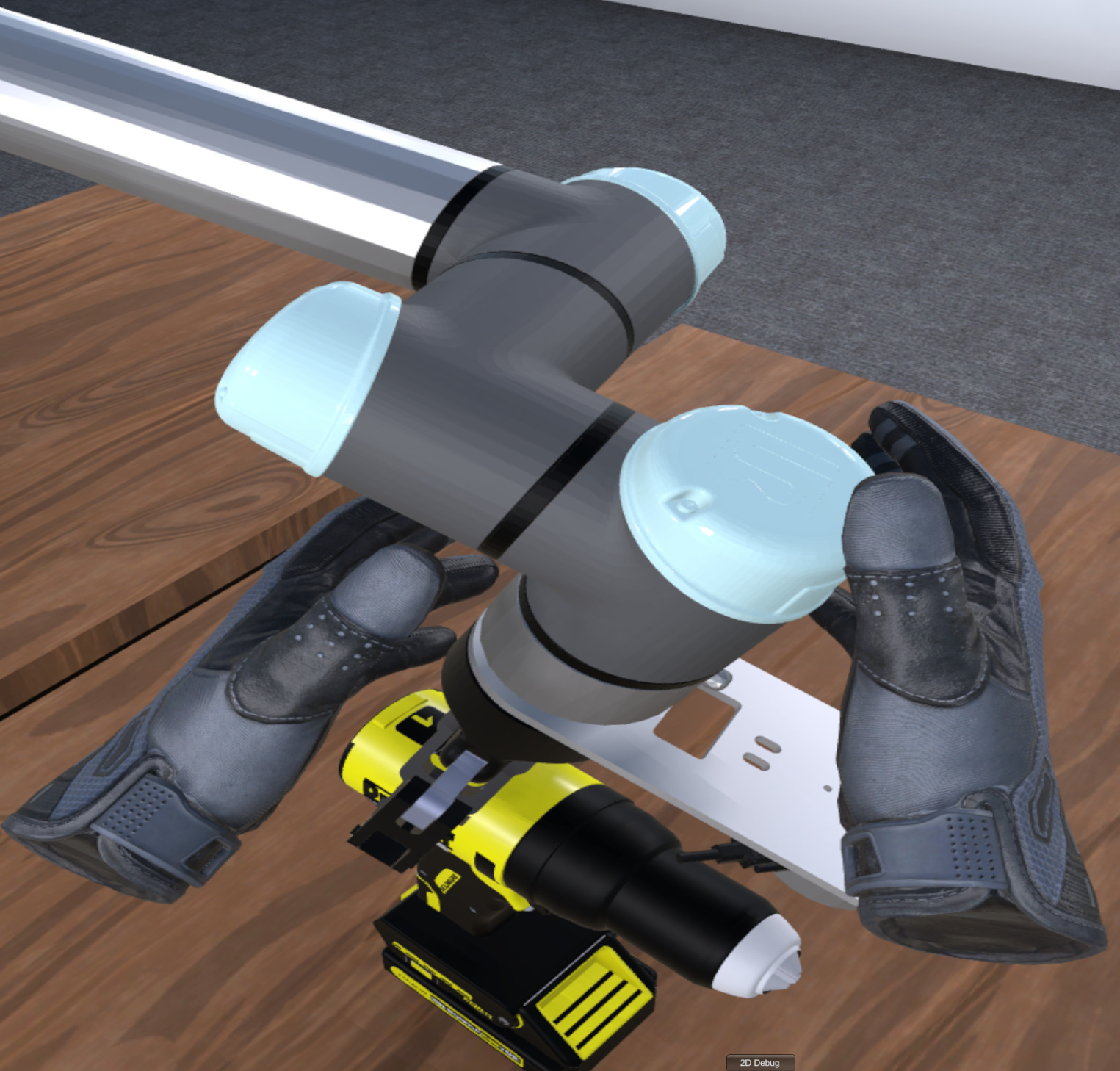}}\
\subfloat[Ghost-drive mode\label{ghostdrive}]{\includegraphics[width=0.49\linewidth, height=1.6in]{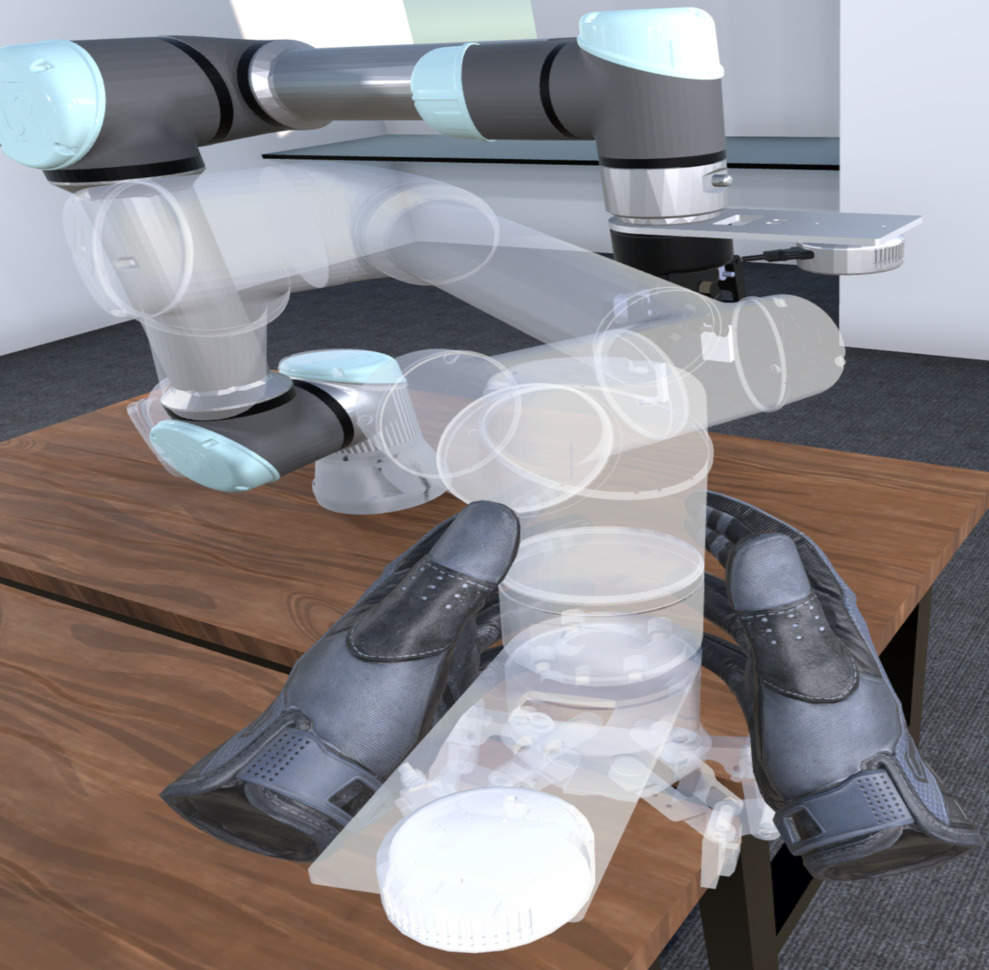}}\hfill
\caption{Different modes of robot operation. (a) Direct-body interaction between
the controlling operator and the body of the virtual robot to pull it along. (b)
Ghost images indicating and recording where the robot will be without actually
moving the body of the robot.}
\label{fig:modes}
\vspace{-1mm}
\end{figure}

We utilize ROS 1 Noetic due to its wide support for a range of industrial robots
from various manufacturers (e.g., Universal Robots, Fetch Robotics, Clearpath
Robotics, etc.), as well as a number of end-effector providers (e.g., Robotiq).
Although, we considered moving to ROS 2 Humble and Rolling for this work, we
ultimately decided to stay with ROS 1 due to incomplete robot support in ROS 2.
ROS 1 Noetic is modern enough to support Python 3, while still providing proper
support for the majority of the robots active in industry. Furthermore, although
ROS 2 is officially supported for Unity packages, it is still under heavy
development and continues to roll out bug fixes and updates.

Motion planning is broken up into the following categories: planning from Unity
and planning from MoveIt. These categories are separate, but they communicate in
between the ROS and Unity backends. Fig.~\ref{fig:modes} shows the point of view
of the operator controlling the simulated robot. From within the Unity visual
scene, the operator can directly drag individual joints into position thus
performing FK or pseudo-FK, whereby they can specify if the robot should follow
along (free-drive mode). Conversely, the operator can overlay a representative
semi-transparent double to indicate the intended final pose (ghost-drive mode).
The operator can also exclusively specify the end-effector pose (position and
orientation) and request IK from MoveIt. 

The underlying motion planning is configurable, commonly employing the Open
Motion Planning Library \cite{sucan2012ompl}, a collection of state-of-the-art
motion planning algorithms. FK is therefore trivial and can be actuated from
within Unity itself, while more difficult motion planning can be delegated to
MoveIt. The position of the robot is constrained by the state provided by the
fake joint controller generated by MoveIt. This prevents discrepancies between
ROS joint states and Unity joint states. Moreover, it is coupled with the robot
representation between Unity and ROS to ensure synchronous behavior. 


\section{Evaluation}
\label{sec:evaluation}
In this section, we showcase the capability of our proposed VRRW to visualize
and train simple robotic tasks. To demonstrate the ability to train a robot in
VR and then transfer the training to a real robot, we utilize a physical robot
setup. The simulation directly mirrors our real workspace and consists of a
Universal Robots (UR) UR5e with a Robotiq 2F-85 2-Finger gripper attached to its
tool port. This adds an additional layer of software communication between the
ROS backend, the physical robot, and Unity, which is made possible using our
open-source UR-Robotiq integrated driver \cite{tram2022ur}.

\subsection{Use Cases}
\label{subsec:use_cases}
We present the following use cases for VRRW. In the first use case, we perform
direct imitation learning on a robot solely through virtual interaction for a
pick-and-place task. In the second use case, we generalize the pick-and-place
behavior by learning from a virtual demonstration using a framework called
dynamic movement primitives (DMPs) \cite{ijspeert2013dynamical}.

\subsubsection{Virtual Direct Imitation Learning}
\label{subsubsec:virtual_direct_imitation_learning}
This use case targets offline, exact replay of recorded robot motion provided by
an operator directly manipulating the body of the robot. It should not be
confused with other forms of learning from demonstration where only the initial
motion is recorded and fed to a machine learning algorithm to generate a
different, but behaviorally similar, trajectory. These direct prerecorded robot
motions are often used as part of a larger, collaborative workflow where
individual robots repeatedly perform their assigned task. Therefore, while the
individual instructions may be simple, when enough of them are properly
orchestrated a collection of mutually dependent motions can achieve complex
results.

A standard, high-level definition of the robot instruction set for a
pick-and-place task can be enumerated as follows.
\begin{enumerate}[i)]
  \item Move robot from current position to near-pick position
  \item Move gripper to pick position
  \item Close (or activate) gripper
  \item Return to near-pick position for clearance (optional)
  \item Move to final position
  \item Open (or deactivate) gripper
\end{enumerate}
An example of a system exercising this use case is an industrial assembly line
(e.g., automotive factory). The independent robot tasks can be as simple as
moving an object from one location to another location, i.e., picking and
placing tools or parts. While the individual robot tasks are straightforward,
their end result is an intricately assembled vehicle. Thus, to demonstrate the
efficacy of our workspace in terms of the ability to train rudimentary robot
tasks from VR, we selected the job of programming a pick-and-place instruction.
From within the VRRW, an operator will manually set and move the simulated arm
through a desired trajectory. This recorded instruction will be replayed using
our real, identical setup to pick up a similar object in the VRRW.

\subsubsection{Virtual Learning from Demonstration}
\label{subsubsec:virtual_learning_from_demonstration}
In this use case, we save the interactions between the VR operator and the robot
as trajectories in the Unity environment to facilitate learning from
demonstration \cite{argall2009survey}. To do this, we make use of motor
primitives, i.e., complex sequences of muscle movements that have been theorized
by neuro-biologists to be composed of \textit{building block} movements. DMPs
attempt to present this motor primitive theory within an elegant mathematical
framework represented by a spring-damper system \cite{ijspeert2002learning,
ijspeert2013dynamical}. Concretely, a DMP system is parameterized by the start
and goal locations, desired velocities, and additional forcing terms that can be
appended to perturb the behavior in response to arbitrary stimuli
(e.g., sensed obstacles).

\begin{figure*}
\centering
\subfloat{\includegraphics[clip,trim=69cm 15cm 0cm 20cm,width=0.24\textwidth]{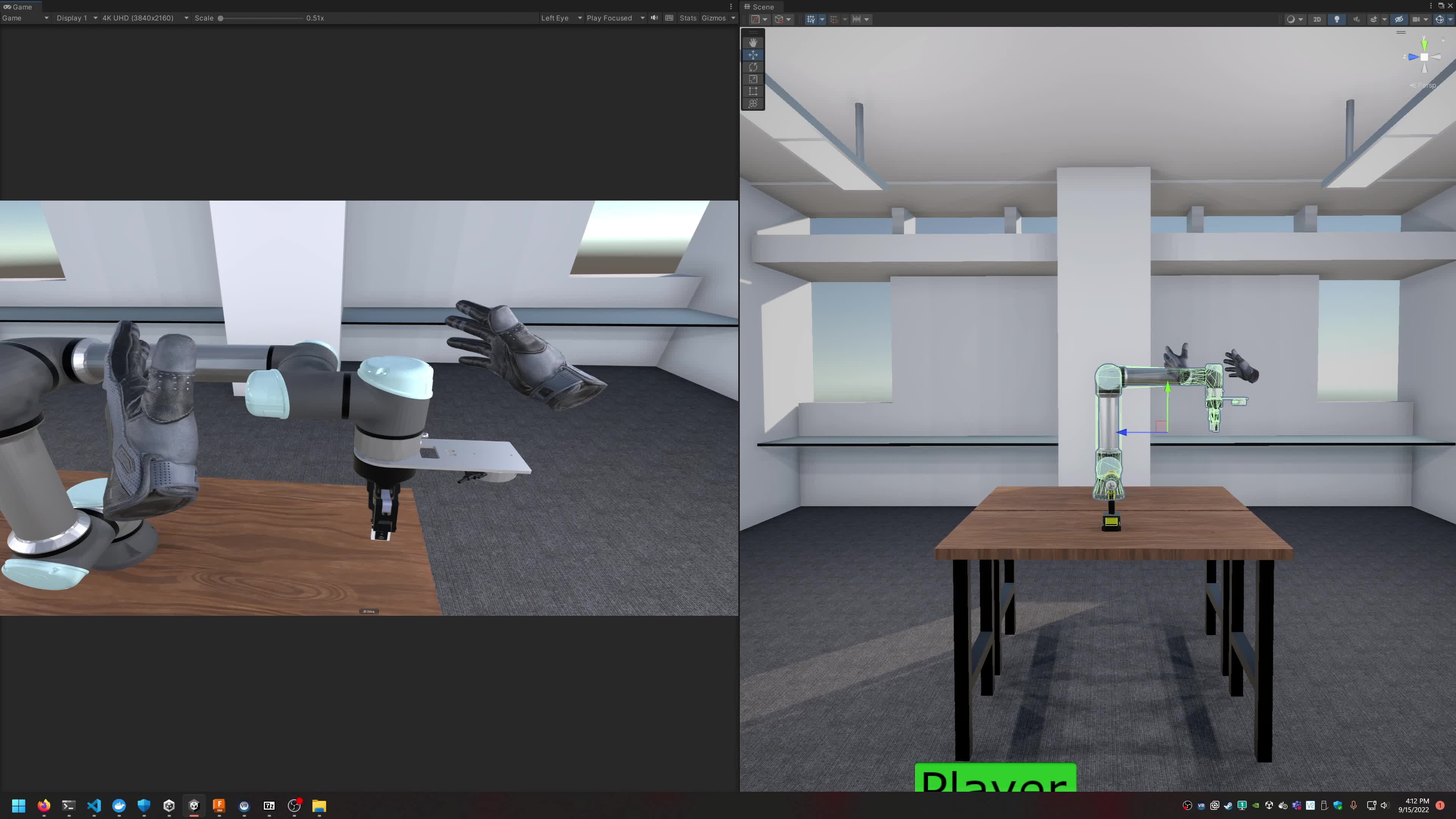}} \enspace
\subfloat{\includegraphics[clip,trim=69cm 15cm 0cm 20cm,width=0.24\textwidth]{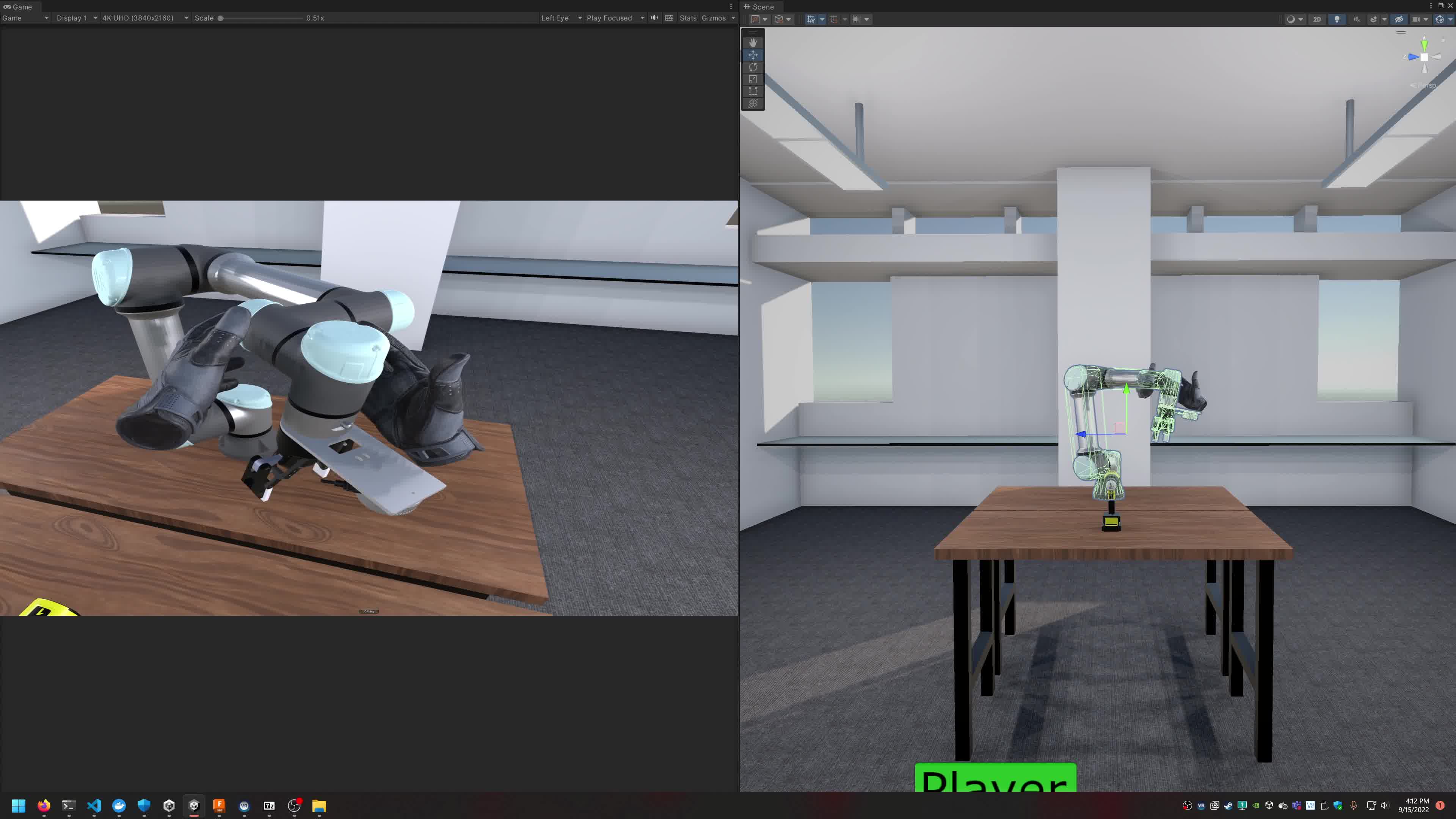}} \enspace
\subfloat{\includegraphics[clip,trim=69cm 15cm 0cm 20cm,width=0.24\textwidth]{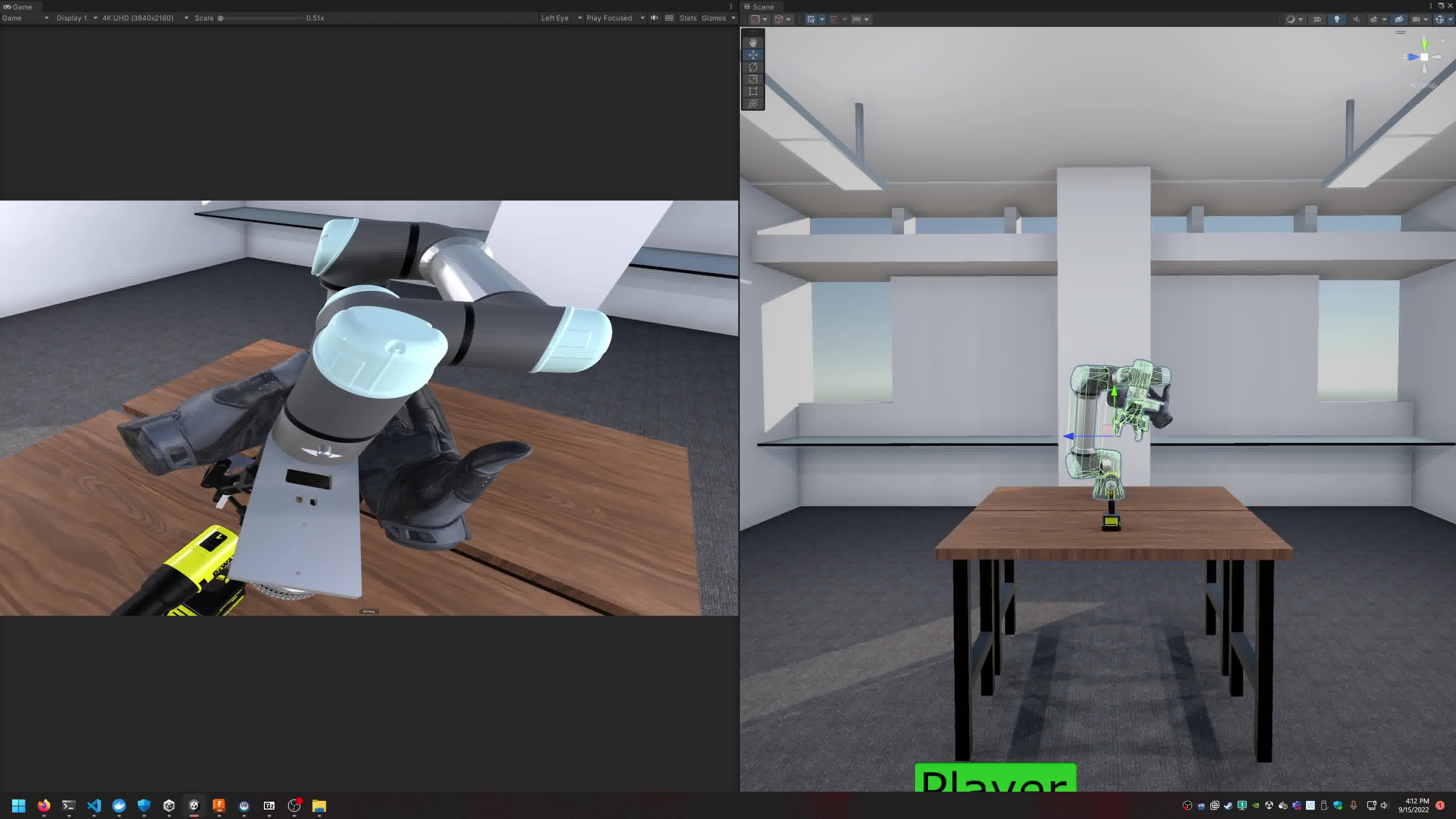}} \enspace
\subfloat{\includegraphics[clip,trim=69cm 15cm 0cm 20cm,width=0.24\textwidth]{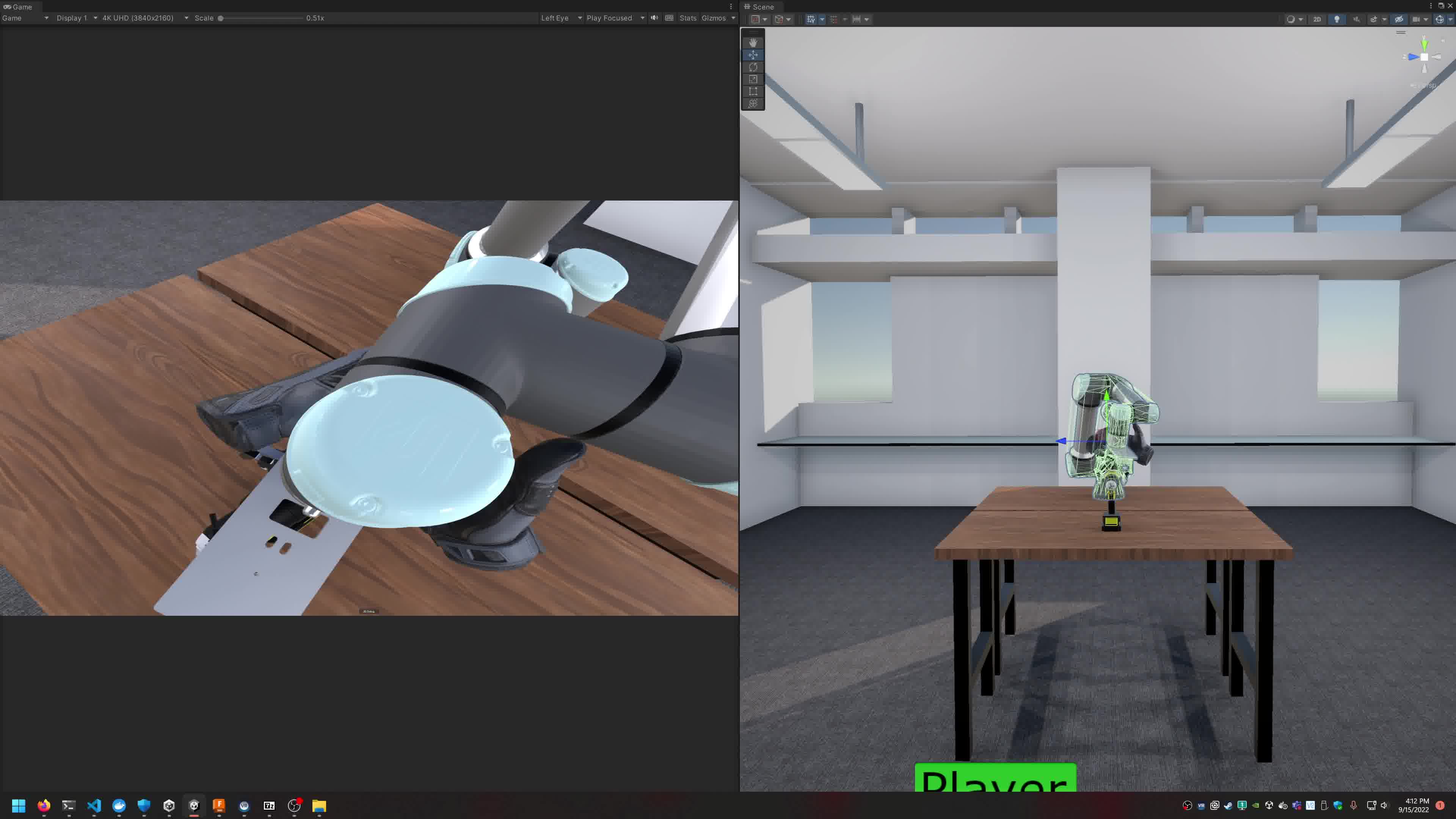}} \hfill
\subfloat{\includegraphics[clip,trim=0cm 0cm 0cm 0cm,width=0.24\textwidth]{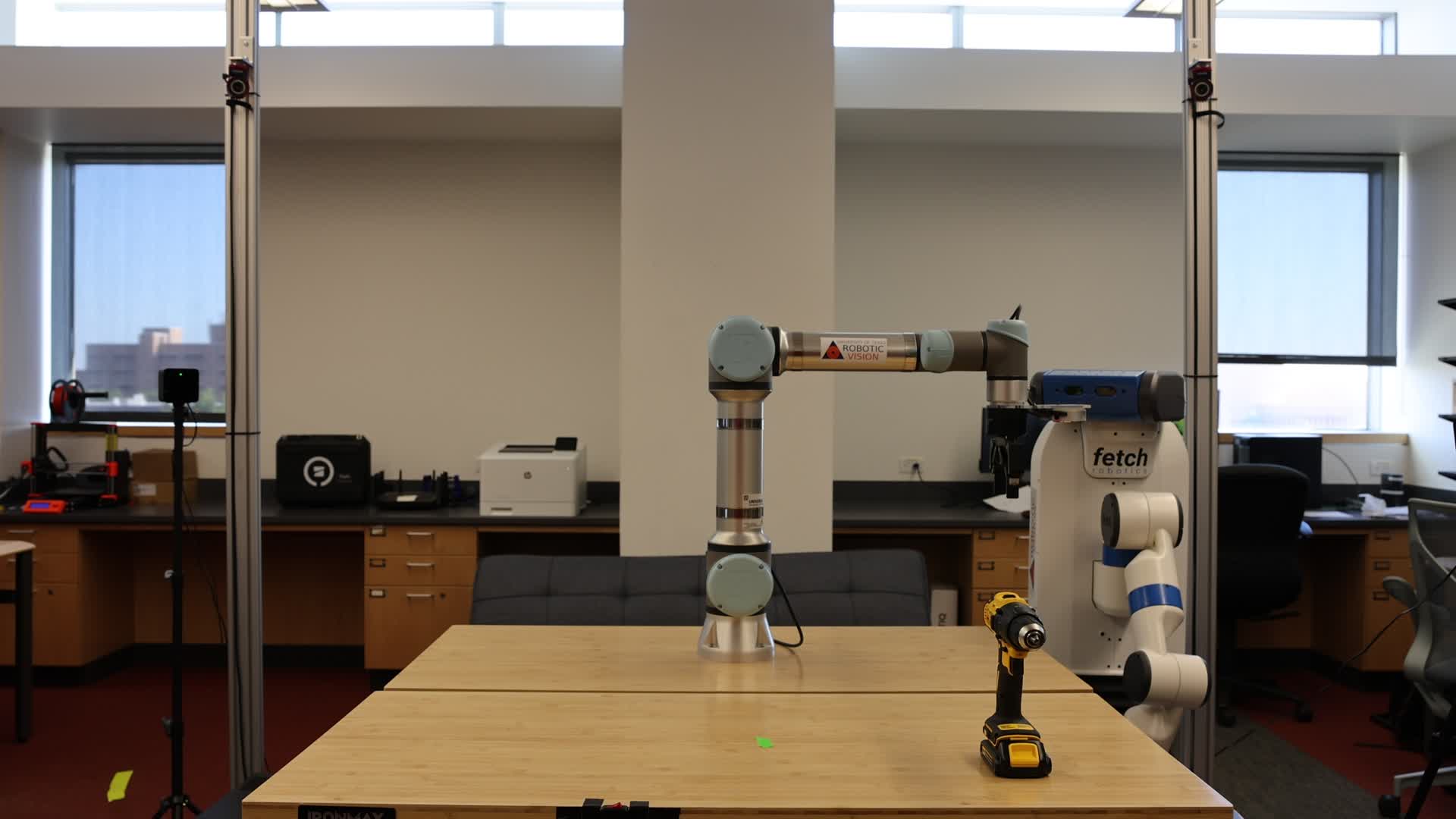}} \enspace
\subfloat{\includegraphics[clip,trim=0cm 0cm 0cm 0cm,width=0.24\textwidth]{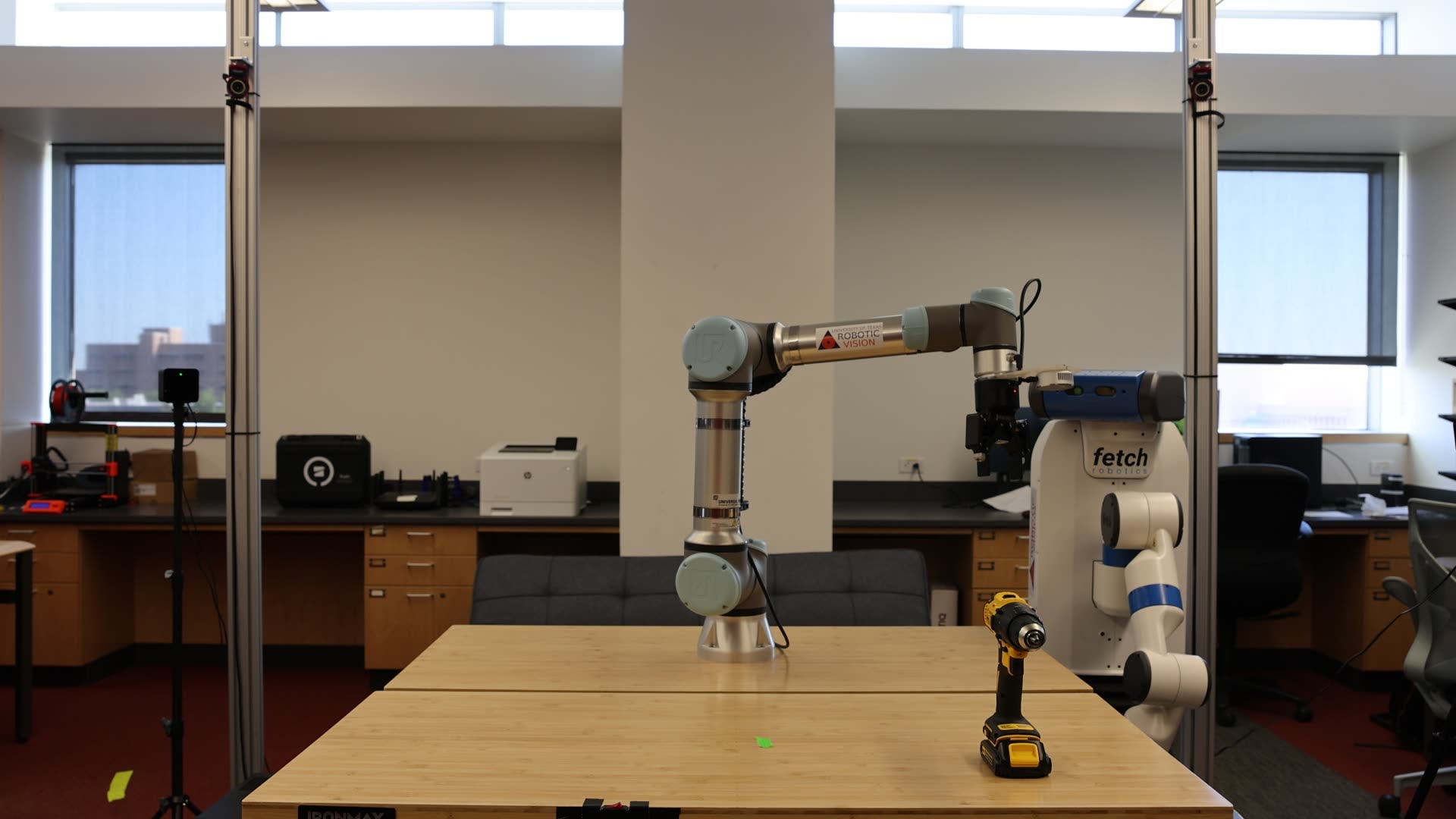}} \enspace
\subfloat{\includegraphics[clip,trim=0cm 0cm 0cm 0cm,width=0.24\textwidth]{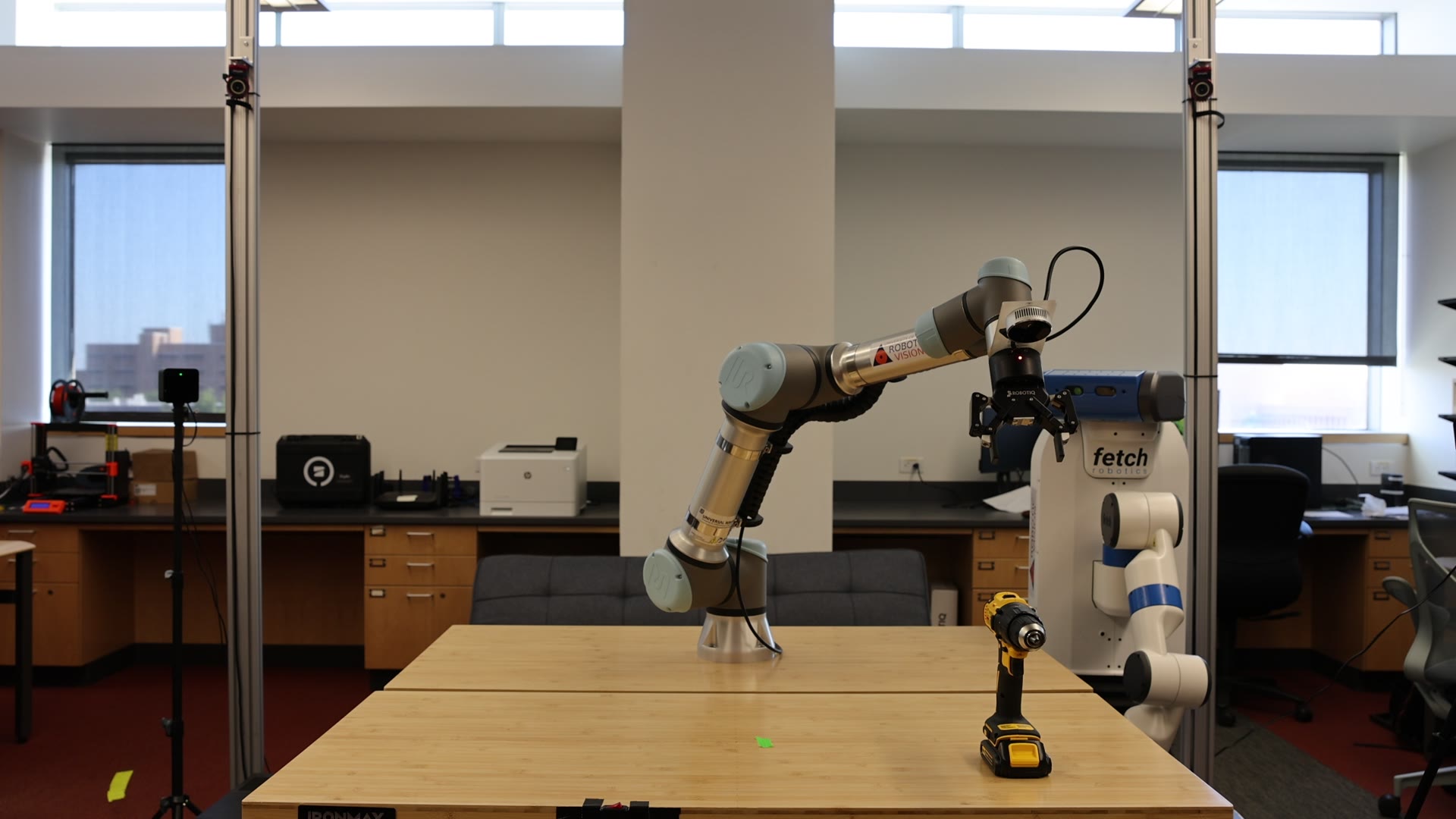}} \enspace
\subfloat{\includegraphics[clip,trim=0cm 0cm 0cm 0cm,width=0.24\textwidth]{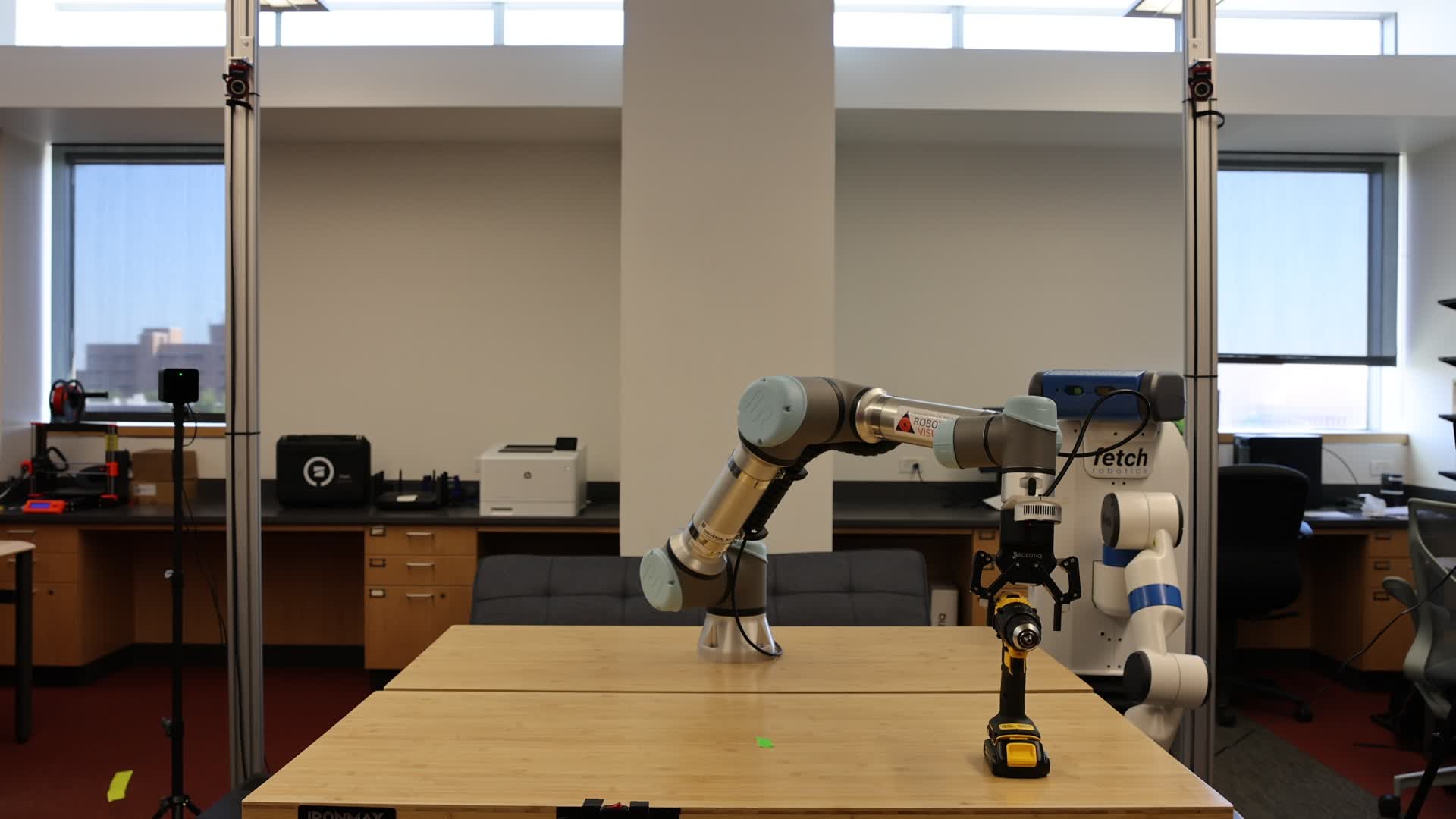}}\hfill
\caption{The temporal execution of the robot pick-and-place task (left to
right). The simulation trajectory was captured using the VRRW (top row). The
second row depicts the DMP execution of the trajectory. Note the novel location
of the object (yellow cordless drill).}
\label{fig:execution_of_simulated_dmp}
\vspace{-4mm}
\end{figure*}

Within our VRRW, we leverage DMPs to learn trajectories demonstrated by the VR
operator. This is done by using the handheld controller to provide the robot
with generalized virtual skills, which can then be executed in both the virtual
and real environments. Formally, a DMP is defined as
\begin{align}
  \tau \dot{v} &= K (g - x) - Dv + (g - x_0)f(s),\nonumber\\
  \tau \dot{x} &= v,
  \label{eq:dmp_old_form_acc}
\end{align}
where $x,v \in \mathbb{R}$ are the position and velocity, $x_0, g \in \RR$ are
the initial and goal positions, and $K,D \in \RR^+$ are constants for the spring
and damping terms. In \eqref{eq:dmp_old_form_acc}, $D = 2 \sqrt{K}$ to render
the system critically damped, $\tau$ is a positive speed scaling factor, and $f$
is the $s$ dependent forcing function to be learned. Exponentially decaying
from 1 to 0, $s$ abstracts away time using the canonical system, i.e.,
\begin{equation}\label{eq:canonical_system}
  \tau \dot{s} = -\alpha s.
\end{equation}
The forcing term is written as
\begin{equation}
  f(s) = \frac{\sum_{i=0}^N \omega_i\,\psi_i(s)}{\sum_{i=0}^N \psi_i(s)}s,
  \label{eq:forcing_term}
\end{equation}
where $N$ Gaussian basis functions, represented by $\psi_i$, are sum weighted
against the learned weights $\omega$. With this, \eqref{eq:dmp_old_form_acc} can
be rewritten to calculate the target forces. Then, the weights can be computed
using locally-weighted regression. Once the weights have been learned, we can
then execute the primitive in novel contexts. We demonstrate this capability on
a pick-and-place task as displayed in Fig.~\ref{fig:execution_of_simulated_dmp}.

\begin{figure*}
\centering
\subfloat[Joint 0
  \label{fig:dmp_joint_0}]{\includegraphics[clip,trim=4cm 2cm 7.5cm 3.5cm,width=0.3\textwidth]{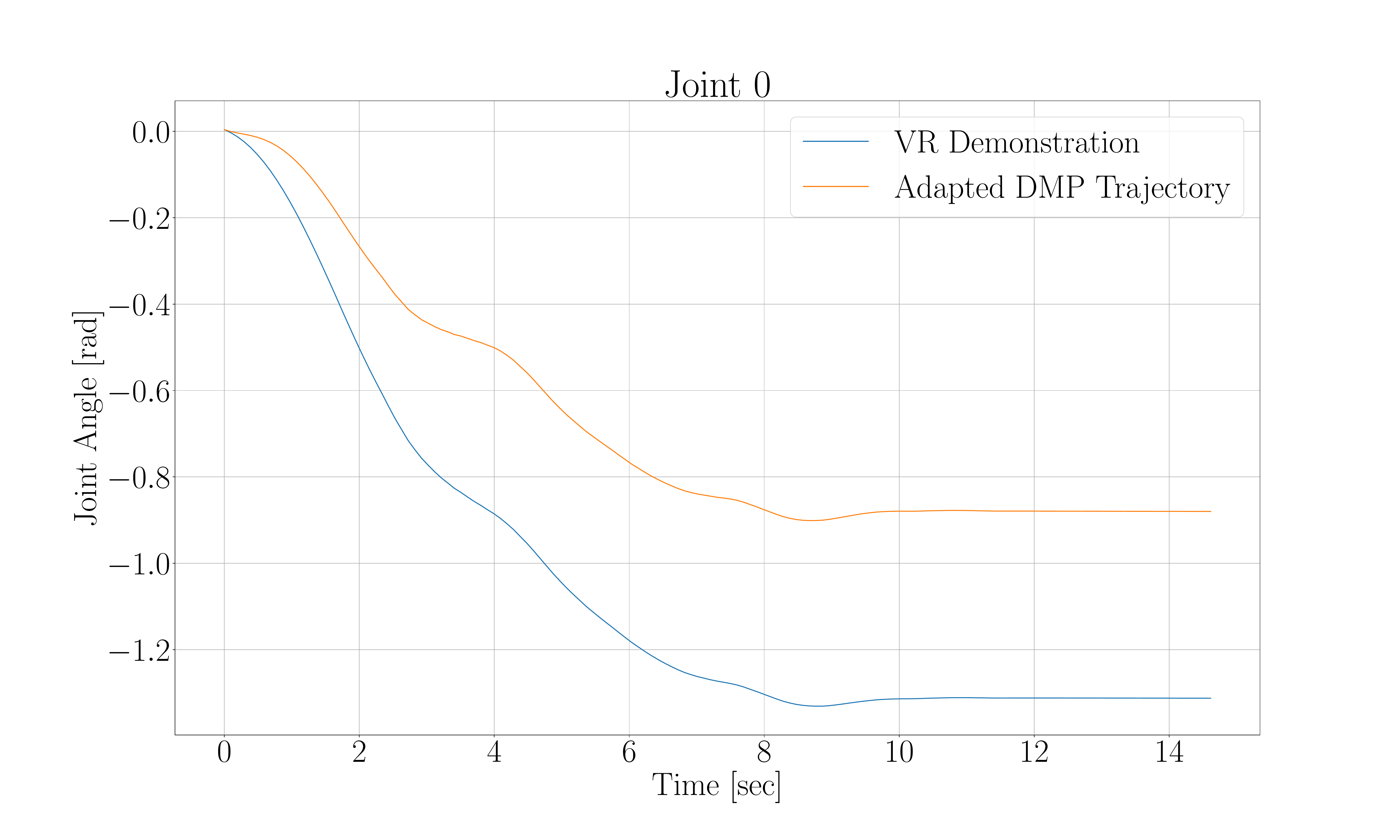}}\qquad
\subfloat[Joint 1
  \label{fig:dmp_joint_1}]{\includegraphics[clip,trim=4cm 2cm 7.5cm 3.5cm,width=0.3\textwidth]{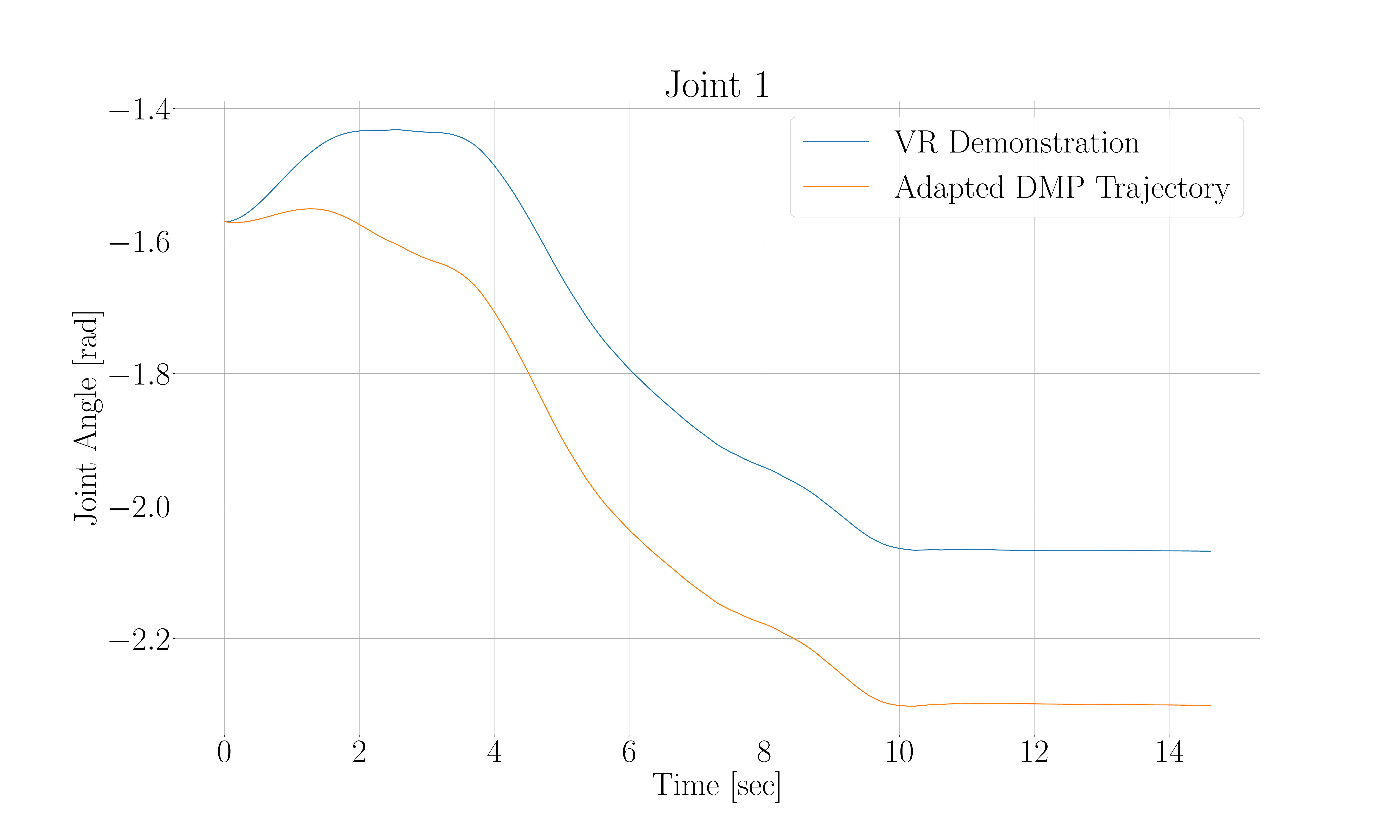}}\qquad
\subfloat[Joint 2
  \label{fig:dmp_joint_2}]{\includegraphics[clip, trim=4cm 2cm 7.5cm 3.5cm,width=0.3\textwidth]{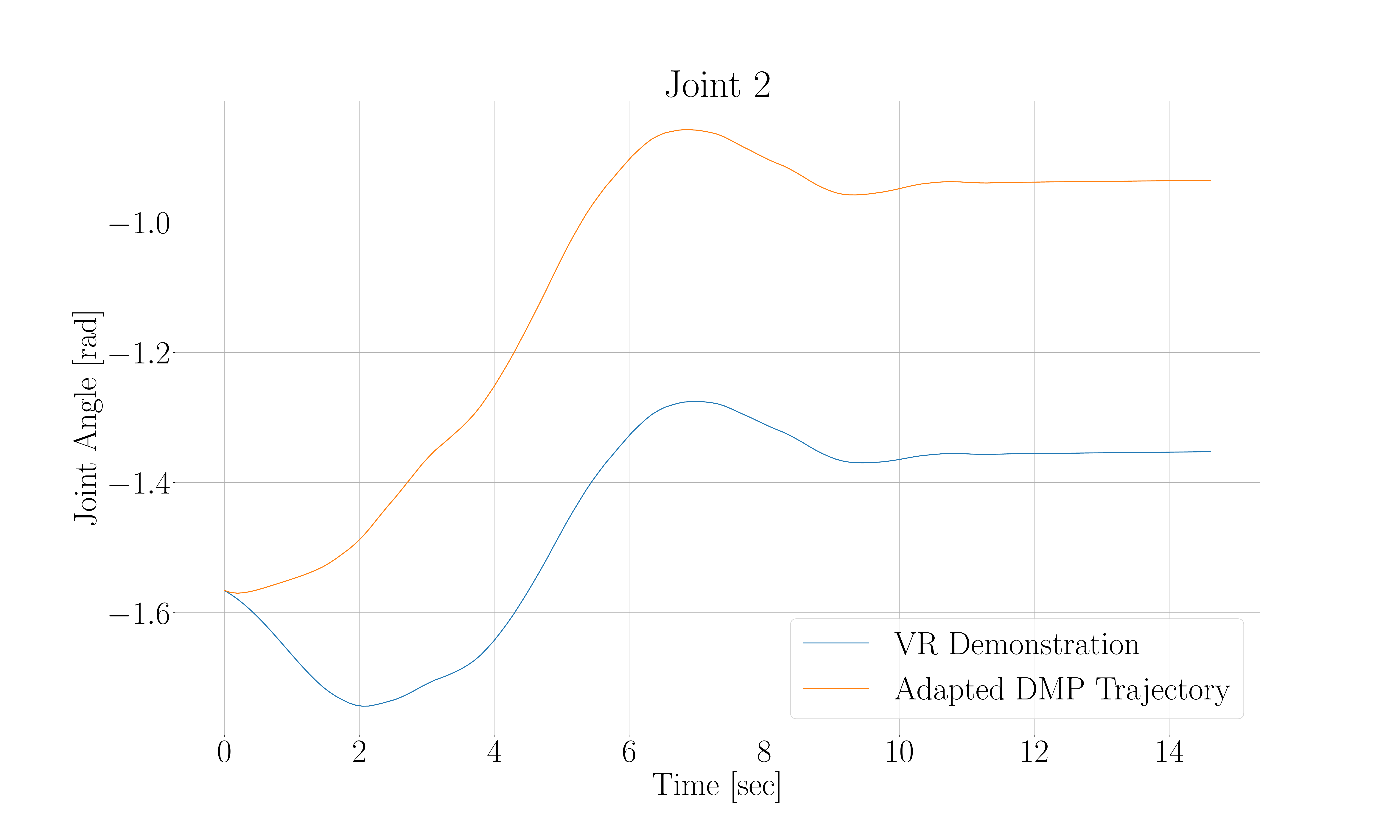}}\hfill
\subfloat[Joint 3
  \label{fig:dmp_joint_3}]{\includegraphics[clip, trim=4cm 2cm 7.5cm 3.5cm,width=0.3\textwidth]{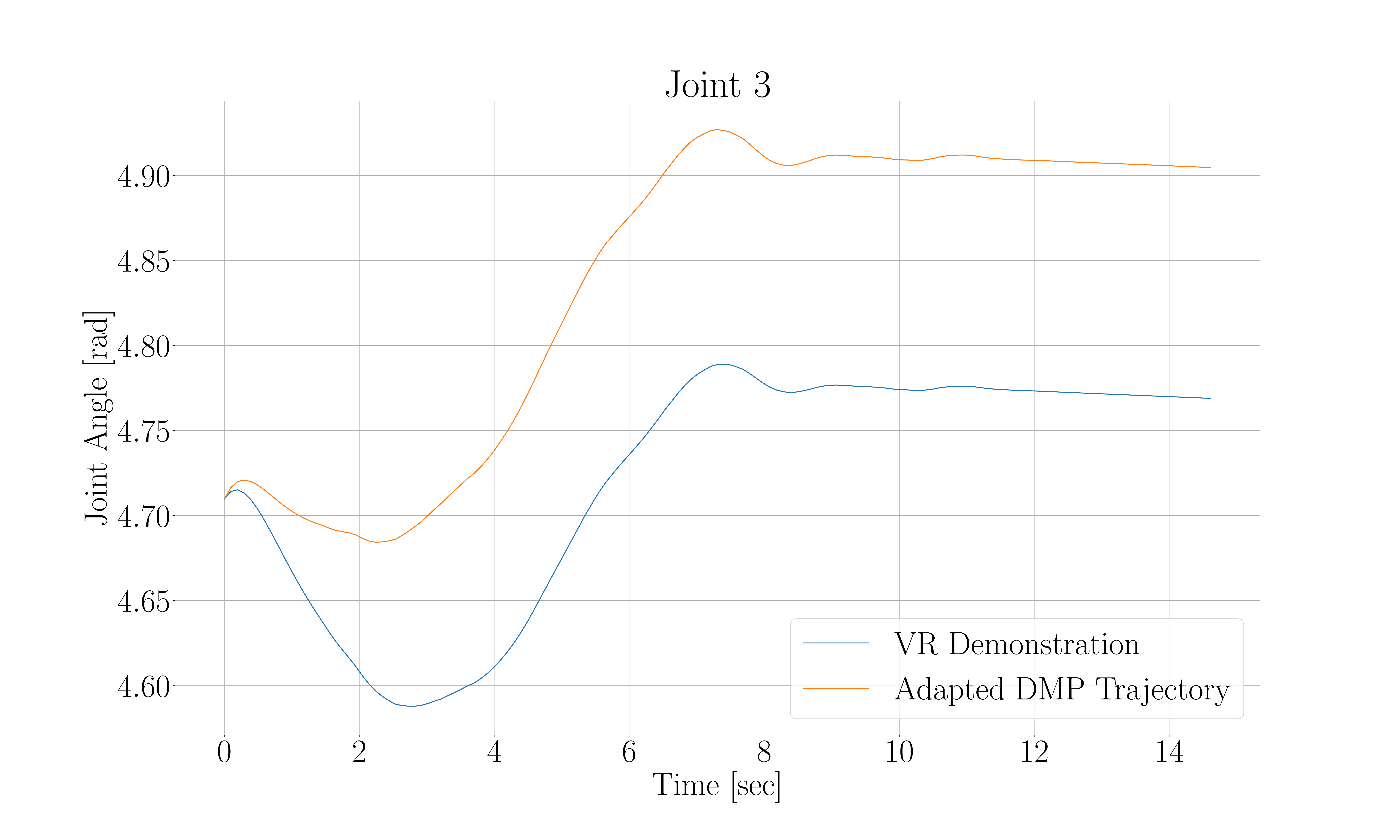}}\hfill
\subfloat[Joint 4
  \label{fig:dmp_joint_4}]{\includegraphics[clip, trim=4cm 2cm 7.5cm 3.5cm,width=0.3\textwidth]{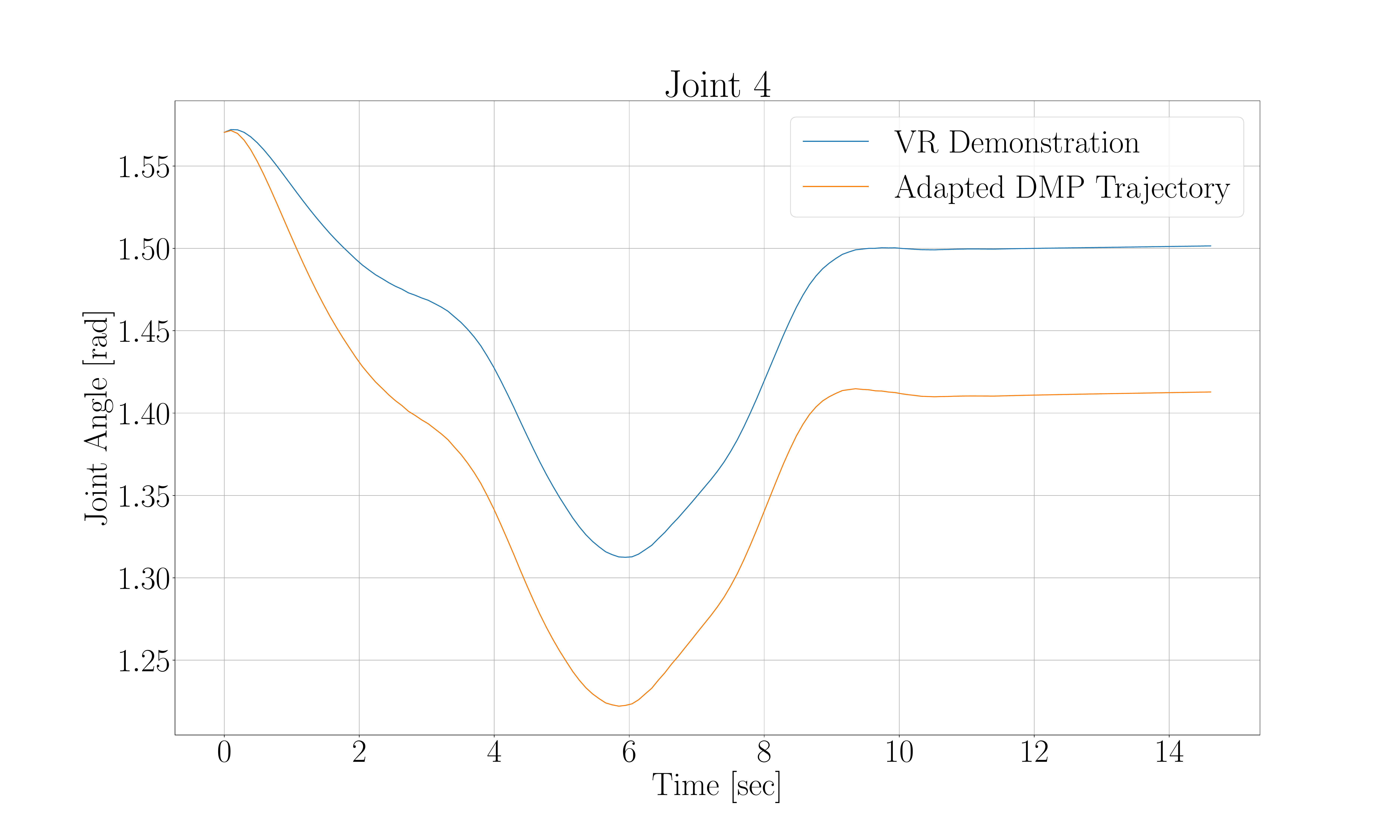}}\hfill
\subfloat[Joint 5
  \label{fig:dmp_joint_5}]{\includegraphics[clip, trim=4cm 2cm 7.5cm 3.5cm,width=0.3\textwidth]{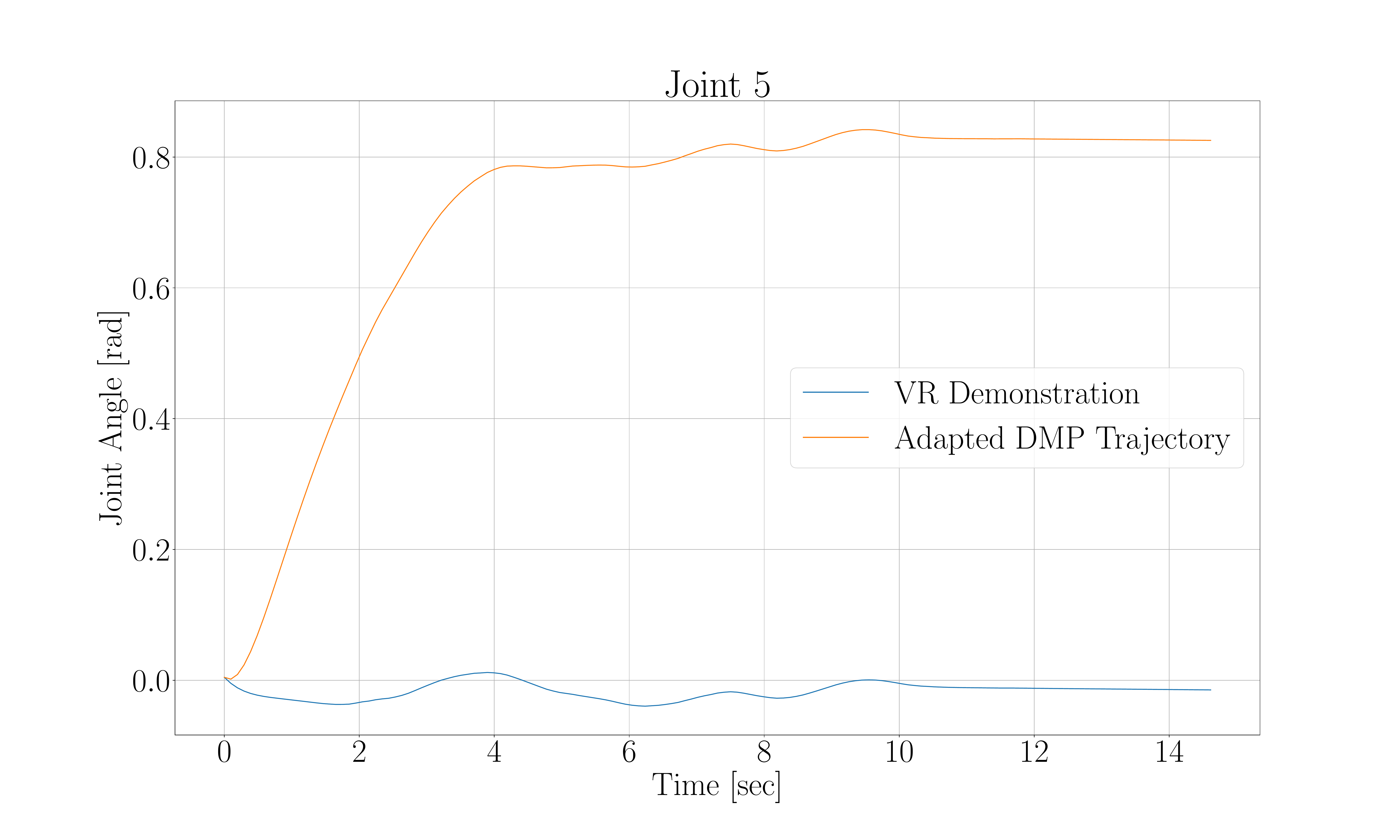}}\hfill
\caption{The UR5e joint trajectories for the virtually trained DMP.}
\label{fig:dmp_trajectories_jointspace}
\vspace{-4mm}
\end{figure*}

\subsection{Discussion}
\label{subsec:discussion}
In the pick-and-place task, we focused on the motion of the robot itself rather
than the actuation of the gripper. We were able to manipulate the simulated
robot directly using natural hand motions (e.g., holding or pushing) on specific
parts of the robot. The trajectory of the robot was recorded as a function of
time for each subtask. We then directly executed the recorded trajectories on a
physically identical robot. Fig.~\ref{fig:dmp_trajectories_jointspace} shows the
correspondence and dynamic adjustments between the virtually recorded and
DMP-generated trajectories. 

Being able to record robot trajectories and other relevant joint state
information naturally allows us to capture training data for machine learning
purposes. Using the joint state information collected exclusively from our VRRW,
we were able to provide sufficient training information for a DMP framework to
adapt a model capable of recreating characteristically similar motions with a
variable goal state. The DMP was able to learn from our VR demonstration and
produce an appropriate new trajectory. The general form of the motion was
retained while being executed from different start and end joint states.
Moreover, the execution of the generated trajectory on the real robot exhibited
similar motion and velocity behavior.

To summarize, the results of the physical execution indicates that there was a
proper transition between the simulation recorded joint states and the physical
joint states, even though there was a slight amount of jitter due to the
discrete time step recording of the simulated trajectory. Overall, the motion
characteristics of the pick-and-place demonstration were preserved and the task
was successfully executed without any deviations from the learned trajectory or
object collisions in the workspace. This shows the ability of our system to
provide meaningful robot access and evaluation. Furthermore, it allows operators
to properly interact with and assess robot integration and programming without
the constraints of obtaining or being in proximity of the physical robot.
\vspace{-2mm}

\section{Conclusion and Future Work}
\label{sec:conclusion_and_future_work}
In this paper, we introduced a novel VR framework for easing the burden of robot
integration and programming. Our VRRW is capable of simulating multiple robots
in a visually captivating and intuitively interactable workspace. We described
the architecture of our system and evaluated its effectiveness against various
robot programming scenarios. Moreover, we showed the ability of VRRW to simulate
desired workspaces and their accompanying robots, as well as its consistency in
simulation to reality transference. For future work, we have additional use
cases (e.g., educational outreach, workforce training, etc.) where VRRW can act
as a foundation for providing access to a robotic workspace. In particular, we
aim to further increase the quality of the visualization and interaction
capabilities of our current approach thus allowing for more natural and
dexterous interactions between operators and virtual environments.

\section*{Acknowledgments} 
The authors acknowledge Stanley Black \& Decker for providing the power tools
used to obtain the research results reported within this paper. Joseph M.
Cloud was supported by a National Science Foundation Graduate Research
Fellowships Program grant (\#1746052).

\bibliographystyle{IEEEtran}
\bibliography{IEEEabrv,intuitive_robot_integration_via_virtual_reality_workspaces}

\begin{thebibliography}{10}
\providecommand{\url}[1]{#1}
\csname url@samestyle\endcsname
\providecommand{\newblock}{\relax}
\providecommand{\bibinfo}[2]{#2}
\providecommand{\BIBentrySTDinterwordspacing}{\spaceskip=0pt\relax}
\providecommand{\BIBentryALTinterwordstretchfactor}{4}
\providecommand{\BIBentryALTinterwordspacing}{\spaceskip=\fontdimen2\font plus
\BIBentryALTinterwordstretchfactor\fontdimen3\font minus
  \fontdimen4\font\relax}
\providecommand{\BIBforeignlanguage}[2]{{%
\expandafter\ifx\csname l@#1\endcsname\relax
\typeout{** WARNING: IEEEtran.bst: No hyphenation pattern has been}%
\typeout{** loaded for the language `#1'. Using the pattern for}%
\typeout{** the default language instead.}%
\else
\language=\csname l@#1\endcsname
\fi
#2}}
\providecommand{\BIBdecl}{\relax}
\BIBdecl

\bibitem{shen2021covid}
Y.~Shen, D.~Guo, F.~Long, L.~A. Mateos, H.~Ding, Z.~Xiu, R.~B. Hellman,
  A.~King, S.~Chen, C.~Zhang, and H.~Tan, ``Robots under covid-19 pandemic: A
  comprehensive survey,'' \emph{IEEE Access}, vol.~9, pp. 1590--1615, 2021.

\bibitem{sanneman2021state}
L.~Sanneman, C.~Fourie, and J.~A. Shah, ``The state of industrial robotics:
  Emerging technologies, challenges, and key research directions,''
  \emph{Foundations and Trends in Robotics}, vol.~8, no.~3, pp. 225--306, 2021.

\bibitem{arevalo2020cues}
S.~Ar\'{e}valo~Arboleda, T.~Dierks, F.~R\"{u}cker, and J.~Gerken, ``There's
  more than meets the eye: Enhancing robot control through augmented visual
  cues,'' in \emph{Proceedings of the ACM/IEEE International Conference on
  Human-Robot Interaction}.\hskip 1em plus 0.5em minus 0.4em\relax New York,
  NY, USA: Association for Computing Machinery, 2020, pp. 104--106.

\bibitem{gharaybeh2019teleop}
Z.~Gharaybeh, H.~Chizeck, and A.~Stewart, ``Telerobotic control in virtual
  reality,'' in \emph{Proceedings of the OCEANS Conference \& Exposition},
  2019, pp. 1--8.

\bibitem{wang2021weld}
Q.~Wang, W.~Jiao, P.~Wang, and Y.~Zhang, ``Digital twin for human-robot
  interactive welding and welder behavior analysis,'' \emph{IEEE/CAA Journal of
  Automatica Sinica}, vol.~8, no.~2, pp. 334--343, 2021.

\bibitem{tsokalo2019twins}
I.~A. Tsokalo, D.~Kuss, I.~Kharabet, F.~H.~P. Fitzek, and M.~Reisslein,
  ``Remote robot control with human-in-the-loop over long distances using
  digital twins,'' in \emph{Proceedings of the IEEE Global Communications
  Conference}, 2019, pp. 1--6.

\bibitem{puljiz2019hand}
D.~Puljiz, E.~Stöhr, K.~S. Riesterer, B.~Hein, and T.~Kröger, ``General hand
  guidance framework using microsoft hololens,'' in \emph{Proceedings of the
  IEEE/RSJ International Conference on Intelligent Robots and Systems}, 2019,
  pp. 5185--5190.

\bibitem{zhang2018learning}
T.~Zhang, Z.~McCarthy, O.~Jow, D.~Lee, X.~Chen, K.~Goldberg, and P.~Abbeel,
  ``Deep imitation learning for complex manipulation tasks from virtual reality
  teleoperation,'' in \emph{Proceedings of the IEEE International Conference on
  Robotics and Automation}, 2018, pp. 5628--5635.

\bibitem{liu2018learning}
H.~Liu, Y.~Zhang, W.~Si, X.~Xie, Y.~Zhu, and S.-C. Zhu, ``Interactive robot
  knowledge patching using augmented reality,'' in \emph{Proceedings of the
  IEEE International Conference on Robotics and Automation}, 2018, pp.
  1947--1954.

\bibitem{dyrstad2018teaching}
J.~S. Dyrstad, E.~Ruud~Øye, A.~Stahl, and J.~Reidar~Mathiassen, ``Teaching a
  robot to grasp real fish by imitation learning from a human supervisor in
  virtual reality,'' in \emph{Proceedings of the IEEE/RSJ International
  Conference on Intelligent Robots and Systems}, 2018, pp. 7185--7192.

\bibitem{vamr2023}
\url{https://github.com/robotic-vision-lab/Virtual-Reality-Robotic-Workspace}.

\bibitem{mukherjee2022survey}
D.~Mukherjee, K.~Gupta, L.~H. Chang, and H.~Najjaran, ``A survey of robot
  learning strategies for human-robot collaboration in industrial settings,''
  \emph{Robotics and Computer-Integrated Manufacturing}, vol.~73, p. 102231,
  2022.

\bibitem{quigley2009ros}
M.~Quigley, B.~Gerkey, K.~Conley, J.~Faust, T.~Foote, J.~Leibs, E.~Berger,
  R.~Wheeler, and A.~Ng, ``Ros: An open-source robot operating system,'' in
  \emph{Proceedings of the IEEE International Conference on Robotics and
  Automation Workshop on Open Source Software}, vol.~3, no. 3.2, 2009, p.~5.

\bibitem{coleman2014reducing}
D.~Coleman, I.~Sucan, S.~Chitta, and N.~Correll, ``Reducing the barrier to
  entry of complex robotic software: a moveit! case study,'' \emph{arXiv
  preprint arXiv:1404.3785}, 2014.

\bibitem{quintero2018programming}
C.~P. Quintero, S.~Li, M.~K. Pan, W.~P. Chan, H.~Machiel Van~der Loos, and
  E.~Croft, ``Robot programming through augmented trajectories in augmented
  reality,'' in \emph{Proceedings of the IEEE/RSJ International Conference on
  Intelligent Robots and Systems}, 2018, pp. 1838--1844.

\bibitem{perez2019virtual}
L.~Pérez, E.~Diez, R.~Usamentiaga, and D.~F. García, ``Industrial robot
  control and operator training using virtual reality interfaces,''
  \emph{Computers in Industry}, vol. 109, pp. 114--120, 2019.

\bibitem{lambrecht2012spatial}
J.~Lambrecht and J.~Krüger, ``Spatial programming for industrial robots based
  on gestures and augmented reality,'' in \emph{Proceedings of the IEEE/RSJ
  International Conference on Intelligent Robots and Systems}, 2012, pp.
  466--472.

\bibitem{kot2018hololens}
T.~Kot, P.~Novák, and J.~Bajak, ``Using hololens to create a virtual operator
  station for mobile robots,'' in \emph{Proceedings of the International
  Carpathian Control Conference}, 2018, pp. 422--427.

\bibitem{gadre2019mr}
S.~Y. Gadre, E.~Rosen, G.~Chien, E.~Phillips, S.~Tellex, and G.~Konidaris,
  ``End-user robot programming using mixed reality,'' in \emph{Proceedings of
  the IEEE International Conference on Robotics and Automation}, 2019, pp.
  2707--2713.

\bibitem{ostanin2020mr}
M.~Ostanin, S.~Mikhel, A.~Evlampiev, V.~Skvortsova, and A.~Klimchik,
  ``Human-robot interaction for robotic manipulator programming in mixed
  reality,'' in \emph{Proceedings of the IEEE International Conference on
  Robotics and Automation}, 2020, pp. 2805--2811.

\bibitem{ghiringhelli2014multi}
F.~Ghiringhelli, J.~Guzzi, G.~A. Di~Caro, V.~Caglioti, L.~M. Gambardella, and
  A.~Giusti, ``Interactive augmented reality for understanding and analyzing
  multi-robot systems,'' in \emph{Proceedings of the IEEE/RSJ International
  Conference on Intelligent Robots and Systems}, 2014, pp. 1195--1201.

\bibitem{chen2019mobile}
I.~Y.-H. Chen, B.~MacDonald, and B.~Wunsche, ``Mixed reality simulation for
  mobile robots,'' in \emph{Proceedings of the IEEE International Conference on
  Robotics and Automation}, 2009, pp. 232--237.

\bibitem{valveindex}
\BIBentryALTinterwordspacing
\emph{Steam Valve Index Headset}, 2023. [Online]. Available:
  \url{https://www.valvesoftware.com/en/index/headset}
\BIBentrySTDinterwordspacing

\bibitem{unity}
\BIBentryALTinterwordspacing
\emph{Unity 3D Game Engine}, 2023. [Online]. Available:
  \url{https://unity.com/}
\BIBentrySTDinterwordspacing

\bibitem{koenig2004design}
N.~Koenig and A.~Howard, ``Design and use paradigms for gazebo, an open-source
  multi-robot simulator,'' in \emph{Proceedings of the IEEE/RSJ International
  Conference on Intelligent Robots and Systems}, vol.~3, 2004, pp. 2149--2154.

\bibitem{krupke2018multimodal}
D.~Krupke, F.~Steinicke, P.~Lubos, Y.~Jonetzko, M.~Görner, and J.~Zhang,
  ``Comparison of multimodal heading and pointing gestures for co-located mixed
  reality human-robot interaction,'' in \emph{Proceedings of the IEEE/RSJ
  International Conference on Intelligent Robots and Systems}, 2018, pp. 1--9.

\bibitem{bambusek2019spatial}
D.~Bambuŝek, Z.~Materna, M.~Kapinus, V.~Beran, and P.~Smrž, ``Combining
  interactive spatial augmented reality with head-mounted display for end-user
  collaborative robot programming,'' in \emph{Proceedings of the IEEE
  International Conference on Robot and Human Interactive Communication}, 2019,
  pp. 1--8.

\bibitem{physx}
\BIBentryALTinterwordspacing
\emph{NVIDIA PhysX 4.0 Physics Engine SDK}, 2023. [Online]. Available:
  \url{https://developer.nvidia.com/physx-sdk}
\BIBentrySTDinterwordspacing

\bibitem{sucan2012ompl}
I.~A. {\c{S}}ucan, M.~Moll, and L.~E. Kavraki, ``The open motion planning
  library,'' \emph{IEEE Robotics \& Automation Magazine}, vol.~19, no.~4, pp.
  72--82, 2012.

\bibitem{tram2022ur}
\BIBentryALTinterwordspacing
M.~Q. Tram, \emph{UR-Robotiq Integrated Driver}, 2023. [Online]. Available:
  \url{https://github.com/robotic-vision-lab/UR-Robotiq-Integrated-Driver}
\BIBentrySTDinterwordspacing

\bibitem{ijspeert2013dynamical}
A.~J. Ijspeert, J.~Nakanishi, H.~Hoffmann, P.~Pastor, and S.~Schaal,
  ``Dynamical movement primitives: Learning attractor models for motor
  behaviors,'' \emph{Neural Computation}, vol.~25, no.~2, pp. 328--373, 2013.

\bibitem{argall2009survey}
B.~D. Argall, S.~Chernova, M.~Veloso, and B.~Browning, ``A survey of robot
  learning from demonstration,'' \emph{Robotics and Autonomous Systems},
  vol.~57, no.~5, pp. 469--483, 2009.

\bibitem{ijspeert2002learning}
A.~Ijspeert, J.~Nakanishi, and S.~Schaal, ``Learning attractor landscapes for
  learning motor primitives,'' in \emph{Proceedings of the Advances in Neural
  Information Processing Systems}, vol.~15, 2002.

\end{thebibliography}

\end{document}